\documentclass[lettersize,journal]{IEEEtran}
\usepackage{amsmath,amsfonts}
\usepackage{algorithmic}
\usepackage{algorithm}
\usepackage{array}
\usepackage[caption=false,font=normalsize,labelfont=sf,textfont=sf]{subfig}
\usepackage{textcomp}
\usepackage{stfloats}
\usepackage{url}
\usepackage{verbatim}
\usepackage{graphicx}
\usepackage{cite}
\usepackage{amssymb}
\usepackage{booktabs}
\usepackage{multirow}
\usepackage{xspace}
\newcommand{\ie}{{\emph{i.e.}}\xspace}

\newcommand{\eg}{{\emph{e.g.}}\xspace}
\newcommand{\etal}{{\emph{et al.}}\xspace}

\usepackage{makecell}
\usepackage{multirow}
\usepackage{bbding}
\usepackage{pifont}
\usepackage{float}
\usepackage{wasysym}
\usepackage{utfsym}
\usepackage{arydshln}

\usepackage[pagebackref=true,breaklinks=true,letterpaper=true,colorlinks,bookmarks=false]{hyperref}

\begin{document}

\title{\emph{ADPS}: Asymmetric Distillation Post-Segmentation for Image Anomaly Detection}

\author{Peng~Xing, Hao~Tang, Jinhui~Tang,~\IEEEmembership{Senior Member,~IEEE,} and~Zechao Li,~\IEEEmembership{Senior Member,~IEEE}
}

\markboth{Journal of \LaTeX\ Class Files,~Vol.~14, No.~8, August~2021}%
{Shell \MakeLowercase{\textit{et al.}}: A Sample Article Using IEEEtran.cls for IEEE Journals}


\maketitle

\begin{abstract}

    Knowledge Distillation-based Anomaly Detection (KDAD) methods rely on the teacher-student paradigm to detect and segment anomalous regions by contrasting the unique features extracted by both networks. However, existing KDAD methods suffer from two main limitations: 1) the student network can effortlessly replicate the teacher network's representations, and 2) the features of the teacher network serve solely as a ``reference standard" and are not fully leveraged. Toward this end, we depart from the established paradigm and instead propose an innovative approach called Asymmetric Distillation Post-Segmentation (ADPS). Our ADPS employs an asymmetric distillation paradigm that takes distinct forms of the same image as the input of the teacher-student networks, driving the student network to learn discriminating representations for anomalous regions.
    Meanwhile, a customized Weight Mask Block (WMB) is proposed to generate a coarse anomaly localization mask that transfers the distilled knowledge acquired from the asymmetric paradigm to the teacher network. Equipped with WMB, the proposed Post-Segmentation Module (PSM) is able to effectively detect and  segment abnormal regions with fine structures and clear boundaries. Experimental results demonstrate that the proposed ADPS outperforms the state-of-the-art methods in detecting and segmenting anomalies. Surprisingly, ADPS significantly improves Average Precision (AP) metric by $\mathbf{9}\%$ and $\mathbf{20}\%$ on the MVTec AD and KolektorSDD2 datasets, respectively. 

\end{abstract}
\begin{IEEEkeywords}
Anomaly detection, Asymmetric distillation, Weight mask block, Post-Segmentation module
\end{IEEEkeywords}

\section{Introduction}

Anomaly detection (AD) strives to identify abnormal images and segment the anomalous regions~\cite{salehi2021unified,pang2021deep,DBLP:journals/tnn/LiuHLTOL22,bergmann2019mvtec}. Its broad applications span from industrial defect detection~\cite{bergmann2019mvtec,carrera2016defect,defard2021padim,fei2020attribute,rudolph2021same,salehi2021multiresolution,deng2022anomaly} and medical image diagnosis~\cite{schlegl2019f,baur2018deep,DBLP:journals/tnn/JuniorY21,mejia2017pca,venkataramanan2020attention} to the emerging domain of autonomous driving~\cite{vojir2021road,bogdoll2022anomaly}. 
However, the task of AD is fundamentally different from general classification or segmentation tasks, as anomalies are hard to define and impossible to categorize exhaustively. Therefore, AD models are trained with normal images. Relying only on normal sample modeling poses a great challenge to detect and segment anomalies.

\begin{figure}
	\centering
	\includegraphics[width=1\linewidth]{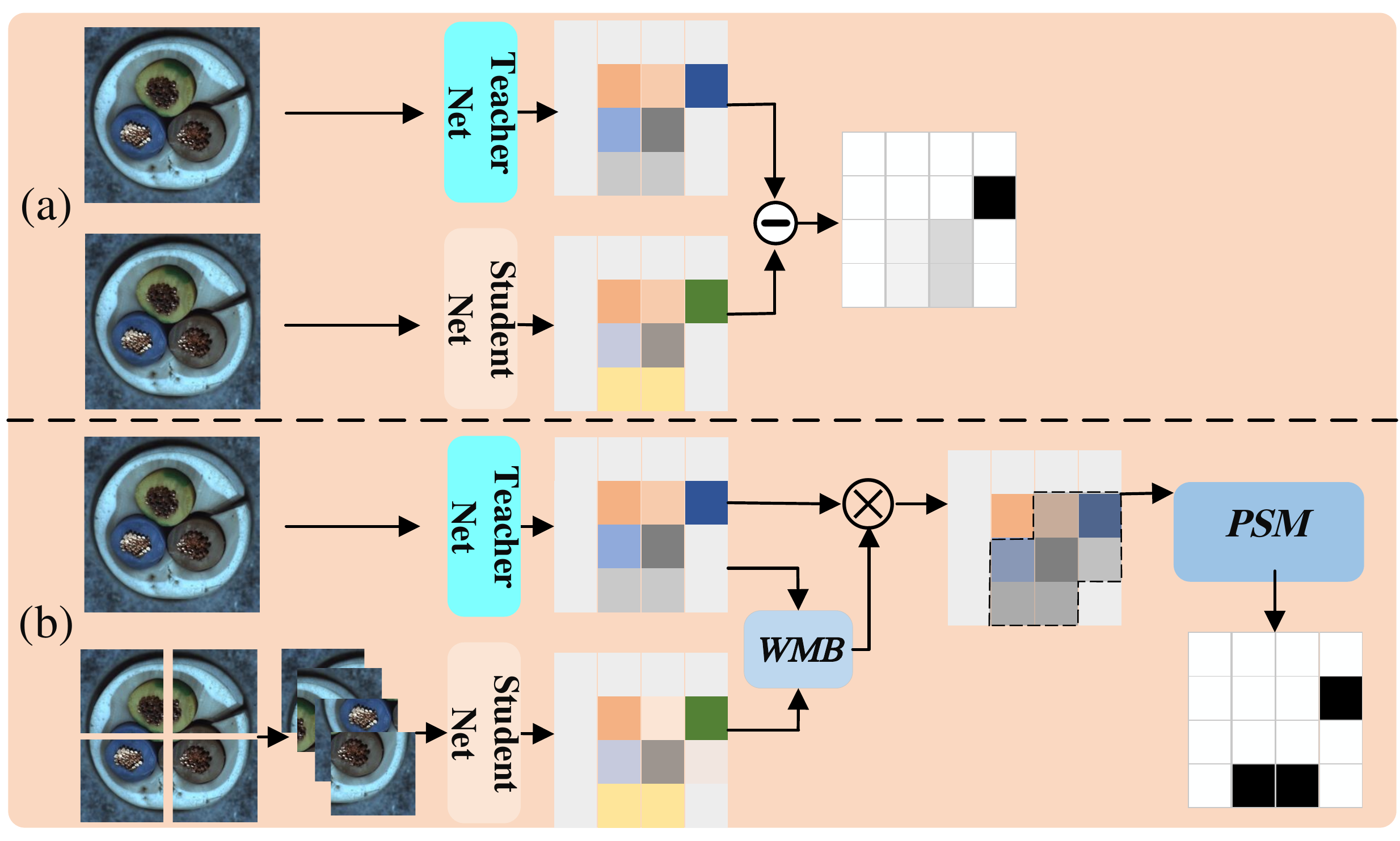}\vspace{-1mm}
	\caption{
 Schematic of the conventional distillation-based symmetric paradigm (a) and the proposed asymmetric paradigm (b).
 } 
 \vspace{-1mm}
	\label{fig:tradvsours}
\end{figure}
Substantial advancements in AD have occurred in recent years, with reconstruction-based methods dominating the landscape~\cite{fei2020attribute,gong2019memorizing,park2020learning,salehi2020puzzle,DBLP:journals/tnn/LiTLW22,DBLP:journals/tnn/ZhouSZLZL22}. The methodology behind these approaches is based on training an autoencoder under a self-supervised paradigm to ensure that out-of-distribution anomalous images are reconstructed with notable errors. The anomalies can then be spotted by assessing the difference between the reconstructed and original image at a high resolution.
%
Innovations such as the self-supervised tasks proposed by Fei \etal \cite{fei2020attribute}, focusing on rotation and color prediction, and by Salehi \etal \cite{salehi2020puzzle}, utilizing image patch sorting, have significantly bolstered the efficacy of this approach.  
Meanwhile, MemAE \cite{gong2019memorizing} undermines the generalization ability of the autoencoder by introducing a memory module that enables the recovery of images solely via the stored features. These techniques aspire to inflate the reconstruction error of the anomalous images by increasing the image reconstruction complexity, thereby proving an effective tool to detect out-of-distribution data ``unseen" by the model.

\begin{figure}
	\centering
	\includegraphics[width=1.0\linewidth]{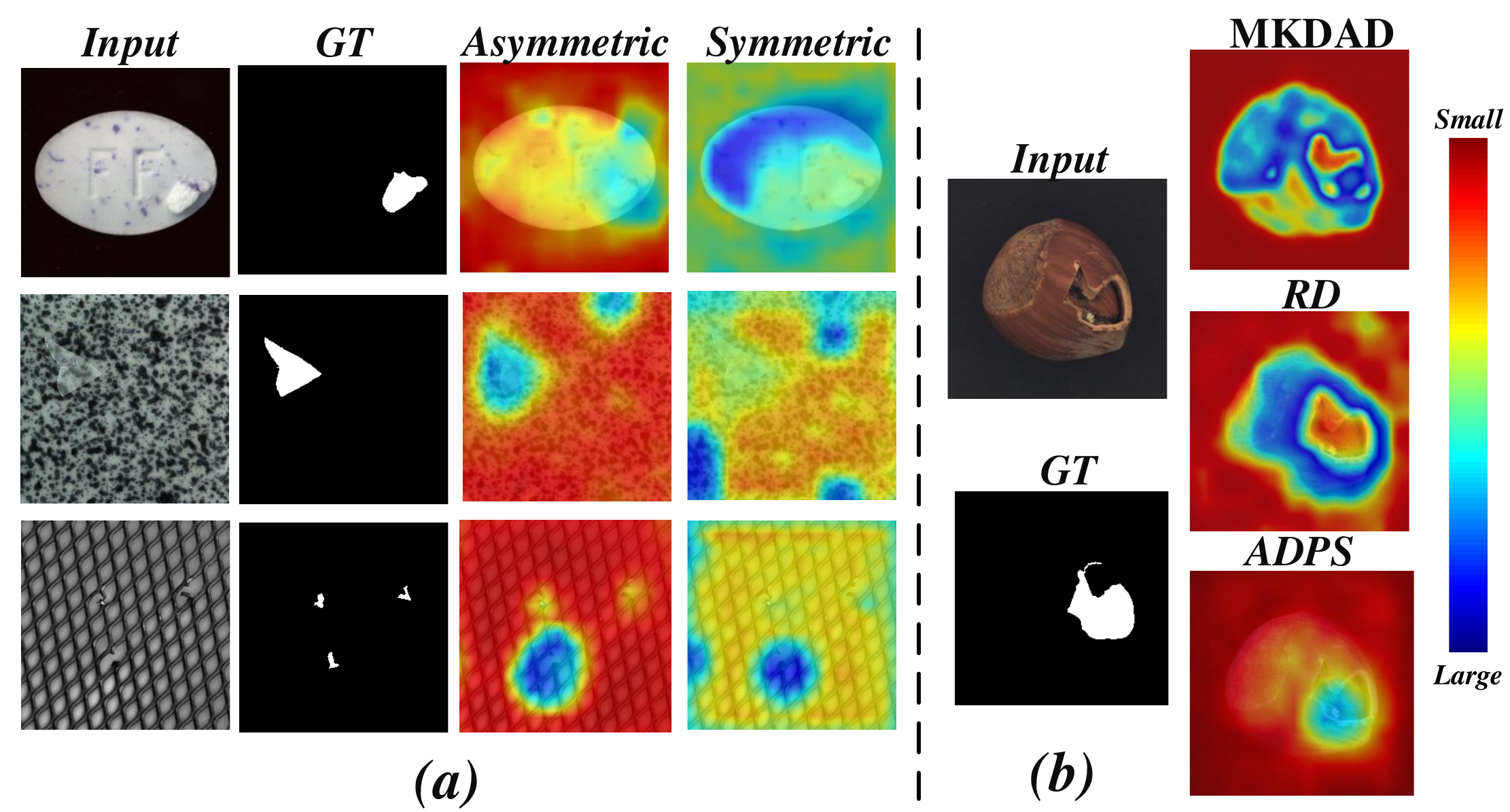}\vspace{-1mm}
	\caption{Visualization of the feature differences extracted by different teacher-student distillation structures. (a) Comparison of using symmetric or asymmetric distillation paradigm in the proposed ADPS. (b) Comparison of ADPS with two representative distillation-based methods, MKDAD \cite{salehi2021multiresolution} and RD \cite{deng2022anomaly}.} 
	\label{fig:figure2}
 \vspace{-1mm}
\end{figure}

Recent studies have shown that knowledge distillation-based anomaly detection (KDAD) methods are promising  \cite{bergmann2019mvtec,bergmann2020uninformed,salehi2021multiresolution,deng2022anomaly, QianWYHW22}, which enable the student network to learn the consistent representation from the teacher network for normal images. Subsequently, the difference in knowledge representation between the teacher network and the student network is utilized as an anomaly detection way, \ie, the difference between the features extracted by the teacher-student networks. In the conventional symmetric distillation paradigm, the teacher network's features are simply used as a ``reference standard" and the regions in which the features extracted by the student network do not match them are identified as anomalies, as shown in Figure \ref{fig:tradvsours}(a). However, existing advanced methods, such as MKDAD \cite{salehi2021multiresolution} and STPM \cite{wang2021student}, limit themselves by not delving into the rich knowledge inherent in the pre-trained teacher network. Furthermore, they stick to a symmetric distillation paradigm which often leads to over-simulation in the student network, \ie, the student model will extract similar representations to the teacher model when confronted with anomalous regions, thereby impeding its ability to generate differentiated representations for anomalous regions.
%
By observing the results of the symmetric distillation models as shown in Figure \ref{fig:figure2}, we experimentally find that the features generated by the anomaly regions only have slight differences compared to those generated by the normal regions, this is particularly problematic when the anomalous regions are similar to the normal regions.

To address the aforementioned issues, we propose a novel \emph{Asymmetric Distillation Post-Segmentation} method, namely ADPS, for image anomaly detection, which departs from the common symmetric distillation paradigm by proposing instead to establish a novel asymmetric distillation paradigm. It is inspired by the prior reconstruction-based autoencoder that utilizes asymmetry to augment the complexity of the student network's alignment with the teacher network. As illustrated in Figure~\ref{fig:tradvsours}(b), ADPS assembles a teacher-student network with a Weight Mask Block (WMB) and a Post-Segmentation Module (PSM).

Firstly, the proposed asymmetric distillation paradigm is applied by ADPS to the teacher-student networks, allowing the same layer of both networks to process diverse forms of the same data. Specifically, the holistic image is assigned to the teacher network while the student network is given non-overlapping patches.
The insight behind the employed asymmetric inputs is that the student network is encouraged to explore unique expressive abilities different from the teacher network, thus enabling the teacher-student networks to generate more differentiated features when meeting anomalous samples (see Figure~\ref{fig:figure2}). Meanwhile, smaller local patches can induce the distillation model to focus on identifying minor anomalies within a local region. 
%
%
Based on the characteristic of the asymmetric distillation paradigm, a customized WMB is proposed to transfer the distilled knowledge to the teacher network, exploiting feature correlations to generate a coarse localization mask. This is expected to improve the discriminability of the teacher model to identify abnormal regions.
%
Contemplating that distilled knowledge may contribute to reducing spurious error and predicting fine anomalies, we further develop a novel PSM that implicitly injects the distilled knowledge into the decoding process to further explore normal/abnormal distributions and obtain high-confidence segmentation results with clear boundaries.

Extensive experiments on three benchmark datasets demonstrate the effectiveness of our ADPS, which outperforms the recent state-of-the-art methods on both the challenging MVTec AD dataset \cite{bergmann2019mvtec} and KolektorSDD2 \cite{bovzivc2021mixed}, achieving more than $9$\% and $20$\% improvement, respectively, in terms of the AP metric for anomaly segmentation.
The main contributions of this work are summarized as follows:

\begin{itemize}
         \item[(1)] We propose a novel asymmetric distillation post-segmentation method. \emph{i.e.,~ADPS}, which establishes an efficient asymmetric knowledge distillation paradigm to boost the performance of image anomaly detection.
         \item[(2)] We carefully design the weight mask block~(WMB) and post-segmentation module~(PSM) to incorporate distilled knowledge and improve the representation of discriminative features for the abnormal region. 
         \item[(3)] Experimental results demonstrate that the proposed ADPS achieves competitive results on three challenging datasets, particularly in solving anomaly detection and anomaly segmentation simultaneously.
\end{itemize}

The remainder of this paper is arranged as follows. Section~\ref{section2} provides a comprehensive discussion about the literature directly associated with our work. The technical details of the proposed ADPS are presented in Section~\ref{section3}. Section~\ref{section4} reports experimental results, ablation studies, and comprehensive analysis. Ultimately, we summarize our findings and suggestions for future research in Section~\ref{section5}.

\section{Related Work}~\label{section2}

\subsection{AutoEncoder-based Anomaly Detection}
AutoEncoder-based methods are a widely used approach for anomaly detection, commonly employing different self-supervised pretext tasks for training and identifying anomalies by measuring the pixel-level difference between the original input and the reconstructed output of the model~\cite{DBLP:conf/visapp/BergmannLFSS19,bogdoll2022anomaly}.
Examples of such methods include AE-SSIM~\cite{DBLP:conf/visapp/BergmannLFSS19}, MemAE~\cite{gong2019memorizing}, MGNAD~\cite{park2020learning}, and DAAD~\cite{hou2021divide}, which mainly employ image reconstruction techniques. Other approaches proposed by Li \etal~\cite{li2020superpixel} and Zavrtanik \etal~\cite{zavrtanik2021reconstruction} utilize image inpainting techniques to complete the content within missing masks, assuming that the model cannot infer anomalous content. However, due to the strong ability of potent pixel-level reconstruction, these AutoEncoder-based models sometimes lead to well-reconstructed anomalies, which is referred to as the ``identity shortcut"~\cite{you2022unified}. To prevent the recovery of abnormal images, 
some studies~\cite{park2020learning,gong2019memorizing,wang2021cognitive,hou2021divide} introduce memory modules that store normal features.  Meanwhile, other methods~\cite{fei2020attribute,salehi2020puzzle} introduce more complex self-supervised tasks, such as coloring and rotation prediction, to detect unseen anomalies that cannot be easily recovered. Despite their strengths, AutoEncoder-based approaches often struggle with anomaly segmentation due to high reconstruction errors in normal regions~\cite{xing2023visual,you2022unified}.

\subsection{Generative Model-based Anomaly Detection}
To overcome the limitations of AutoEncoder-based methods, research has explored Generative Model-based approaches, such as GAN~\cite{DBLP:conf/nips/GoodfellowPMXWOCB14}), as demonstrated in previous works~\cite{schlegl2017unsupervised,akcay2018ganomaly}. In these methods, anomalies are identified if the generated sample significantly differs from the input sample~\cite{anogan/SchleglSWSL17,bogdoll2022anomaly}. AnoGAN~\cite{anogan/SchleglSWSL17} is one such method that employs noise to generate a target image and then compares the differences between the original and the generated image. However, during the inference phase, this model suffers from a significant drawback involving computationally expensive and time-consuming parameter updating processes. To overcome this issue, EffGAN~\cite{EfficientGAN/abs-1802-06222} introduced an encoder that constrains the input noise to its output, successfully alleviating the time-constraint issue. Further refinements in this direction led to GANomaly~\cite{akcay2018ganomaly} and f-AnoGAN~\cite{schlegl2019f}, which improved the generator's ability to impose constraints on spatial feature reconstruction. In parallel, DefGAN~\cite{zhang2020defgan} improved the discriminator segment by integrating multiple discriminators. Recently, Hou~\etal~\cite{hou2021divide} introduced an additional discriminator to distinguish the generated image from the original one, enhancing the quality of generation. While these developments signify substantial progress, they fail to effectively handle anomalous samples with minuscule anomaly regions.

\subsection{Deep Feature Modeling-based Anomaly Detection}
Another distinctive approach to detecting anomalies involves Deep Feature Modelling-based methods, which assume the existence of a gap between normal and abnormal distributions. These methods begin by constructing a feature space specifically for normal images and then determine whether an image is abnormal by evaluating spatial boundaries~\cite{chen2001one,ruff2018deep,chalapathy2018anomaly,yi2020patch,cohen2020sub, defard2021padim}.
Among these methods, SPADE~\cite{cohen2020sub} stands out by deploying a pre-trained WideResNet50~\cite{DBLP:conf/bmvc/ZagoruykoK16} to extract patch features. The presence of anomalies is established by determining whether the tested patch contains $k$ adjacent normal patches. PaDiM~\cite{defard2021padim} builds on this method by further refining the process. It utilizes a pre-trained model for feature extraction and models normality at each location using a multivariate Gaussian distribution~\cite{do2008multivariate,hazel2000multivariate}.
Further efforts have been made to explore the interrelationships within the data. For instance, Liu~\etal~\cite{DBLP:journals/tnn/LiuXLZL23} proposed a method that models the relationship between a data point and its neighbors. Other approaches have attempted to implicitly model the distribution. Rudolph~\etal ~\cite{RudolphWRW22} and Roth~\etal~\cite{RothPZSBG22}, for example, utilize normalizing flow to model the relationship between the original and normal feature distributions. Recent studies~\cite{DBLP:journals/corr/abs-2111-07677, DBLP:journals/corr/abs-2301-12082} similarly seek to model the distribution relationship between patches. Notably, the method presented in~\cite{DBLP:journals/corr/abs-2301-12082} introduces visual rotation invariant features and graph networks. Despite these advancements, a significant challenge remains: these methods have not yet delivered effective results in anomaly segmentation.


\subsection{Knowledge Distillation-based Anomaly Detection}
Knowledge Distillation-based methods leverage discrepancies in the expressive abilities of the teacher-student networks, thereby enabling them to extract differential features when encountering abnormalities. The pioneering work of
U-Std~\cite{bergmann2020uninformed} employed a knowledge distillation model for anomaly detection. MKDAD~\cite{salehi2021multiresolution} was subsequently introduced to address a flaw in U-Std, namely that it only leverages the output of the final layer. Wang~\etal~\cite{wang2021student} proposed the Student-Teacher Feature Pyramid Matching (STPM), implementing the multi-scale feature map approach, which differs from MKDAD by utilizing the discrepancy between multi-feature maps to derive localization maps instead of the gradient of the loss function. However, models with symmetric paradigms or similar structures are unable to extract differentiated features. RD~\cite{deng2022anomaly} employs a pre-trained model as an encoder, enabling the student decoder to align with the encoder. However, it merely employs the output of the teacher network as a ``reference standard" without further exploiting the strong characteristics of the teacher. Additionally, while RD predominantly excels in detecting anomalies, it falls short in accurately segmenting anomalous regions. To address these limitations, ADPS introduces the asymmetric distillation paradigm. This paradigm enhances symmetric distillation paradigms to concentrate on fine anomalies and simultaneously employs  Weighted Mask Block (WMB) and Post-Segmentation Module (PSM) to amalgamate distilled knowledge with pre-trained prior knowledge. This enables a more nuanced exploration of normal and anomalous feature distributions, which is crucial for accurately segmenting fine anomalous regions.
\begin{figure*}
	\centering
	\includegraphics[width=0.99\linewidth]{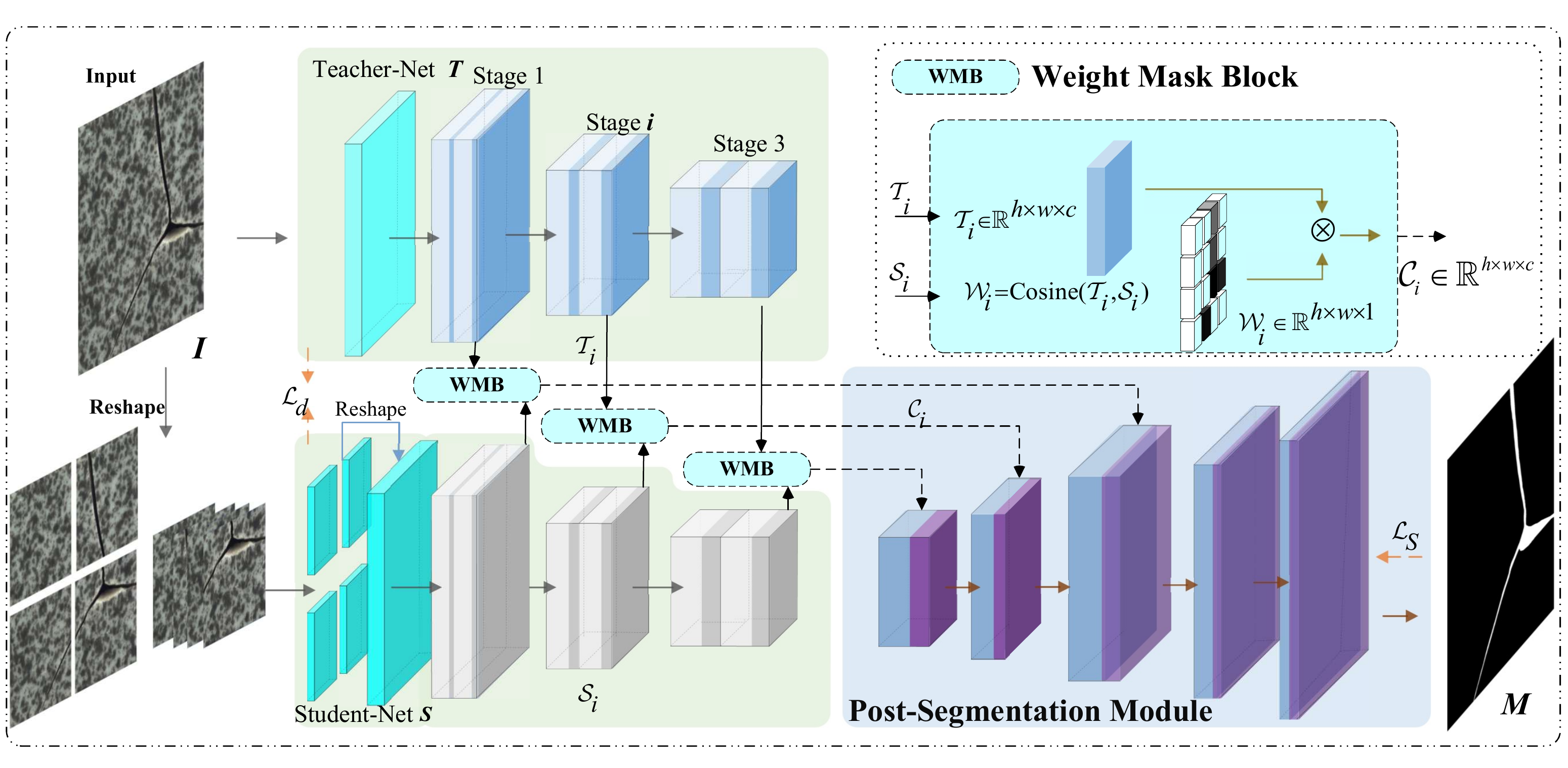}\vspace{-3mm}
	\caption{Overview of the proposed \emph{Asymmetric Distillation Post-Segmentation (ADPS)} framework. The asymmetric distillation paradigm uses asymmetric data (\emph{e.g.,~}$I/I_S$ and $\mathcal{T}_i/\mathcal{S}_i^\prime$) to stimulate $S$ in learning effective discriminative representations. WMB generates a coarse localization mask to transfer distilled knowledge to the teacher model which has a powerful representation capability, leading to the feature $\mathcal{C}_i$. Based on WMB, PSM explicitly learns normal and anomaly feature distribution in 
$\mathcal{C}_i$ with a segmentation module to precisely segment finer anomaly regions.} 
	\label{fig:figureoverview}
 \vspace{-3mm}
\end{figure*}
\section{The Proposed Approach}~\label{section3}

As illustrated in Figure \ref{fig:figureoverview}, this paper proposes a novel method called \emph{Asymmetric Distillation Post-Segmentation (ADPS)}, which includes an asymmetric distillation paradigm, a Weight Mask Block (WMB) and a Post-Segmentation Module (PSM) specialized for anomaly detection. The asymmetric distillation paradigm devises inputs with asymmetry that are utilized to enhance the representation ability gap of the teacher-student networks, thereby generating more discriminative differential features to anomalous regions.
%
The distillation knowledge is further exploited by WMB and injected to the features extracted from the pre-trained teacher network. PSM accepts features of the teacher network consisting of prior knowledge and distilled knowledge as input and utilizes the segmentation module to learn the distribution of normal features. The detailed architecture of our ADPS is described as follows. 

\subsection{Asymmetric Distillation Paradigm}\label{sec3.2}

In the field of Anomaly Detection (AD), 
knowledge distillation has been leveraged to exploit the differential knowledge representation abilities possessed by the teacher and student networks to accurately identify anomalous regions from extracted features. 
However, a crucial drawback of conventional distillation frameworks, such as the symmetric distillation paradigm depicted in Figure \ref{fig:figure2}-(a), 
is that the teacher-student network becomes insensitive to disparities in the representation of anomalous regions. 
This insensitivity stems from the fact that the teacher and student networks tend to learn analogous representations of anomalous regions. Thus, to address these concerns, a feasible solution is to widen the discrepancy in the representational abilities of the teacher-student networks.

The representational abilities of a model are inherently linked to its inputs. Therefore, modifying different inputs for teacher-student networks can obtain different knowledge representation capabilities. We propose an asymmetric distillation paradigm, wherein we construct semantically equivalent, yet structurally different inputs for the teacher-student networks. 
The disparity in format aims to probe their own expressiveness while simultaneously escalating the alignment complexity between the student and teacher networks for unfamiliar anomalous regions. As a result,
the preservation of the same semantic guarantees that, through training, the student network learns the same features as the teacher network for normal samples. Another key factor is that the local features obtained by the student network are affected only by their own patches and do not stray from anomalous feature vectors in remote parts.  

Our ADPS proposes a straightforward yet effective approach for generating distinct but semantically identical data by splitting original images. 
To implement this approach, a given normal image $I \in \mathbb{R}^{H \times W \times C}$,
is split into non-overlapping two-dimensional patches, $I_S =\{I_S^1,...,I_S^j,...,I_S^{k^2}\}$, where $I_S^j \in \mathbb{R}^{ \frac{H}{k} \times \frac{W}{k} \times C}$, and $k^2$ denotes the number of two-dimensional patches. 
In the teacher network, $I$ serves as the direct input, and feature $\mathcal{T}_i$ is obtained by a pre-trained encoder in stage $i$. 
The use of the original image as input in the teacher model assures that $\mathcal{T}_i$ does not lose any information about the original image. 
Meanwhile, in the student network, $I_S$ is utilized as a series of inputs. In fact, this takes $k^2$ forward processes, and in each forward a patch $j$ of $\frac{H}{k} \times \frac{W}{k} \times C$ is fed into the student network to extract $\mathcal{S}_i^j$. During $k^2$ times forward, we can obtain $I$ in the feature representation of the student network, \ie, $\mathcal{S}_i^\prime =\{\mathcal{S}_i^1,...,\mathcal{S}_i^j,...,\mathcal{S}_i^{k^2}\}$. 
During the parameter update, the extracted $\mathcal{S}_i^\prime$ is subsequently reshaped into $\mathcal{S}_i$ based on the original position correspondence, and $\mathcal{S}_i$ is calculated as the loss by distillation loss with $\mathcal{T}_i$.
The approach, named asymmetric distillation paradigm, could theoretically extend to feature distillation in every intermediate layer by simply splitting features extracted by the intermediate layer.

Since the student network receives a patch of $\frac{H}{k} \times \frac{W}{k} \times C$ as the input, it will pay more attention to local regions and detect tiny anomalies. Additionally, the asymmetry between local and global receptive fields can exacerbate the complexity of feature alignment in teacher-student networks. 
%
The student network can acquire unique data representation abilities because of the different inputs.
Therefore, the distilled knowledge of the asymmetric distillation paradigm presents higher discrimination efficiency between normal and abnormal patterns.

\subsection{Weight Mask Block} \label{sec3.3}

Given that prior knowledge in the pre-trained teacher network is acquired from an upstream task, such as image classification, it lacks discriminability for normal and abnormal patterns. 
Towards this end, the proposed Weight Mask Block (WMB) transfers distilled knowledge with the ability to discriminate for anomaly patterns to the teacher network in a manner that promotes the acquisition of features that are effective and beneficial for anomaly detection. By ``distilled knowledge", refers to the coarse localization information obtained through asymmetric distillation for anomalous regions. 
Specifically, features extracted from the teacher and student networks in the asymmetric distillation paradigm are utilized to generate a coarse localization mask. We then take a weighted approach to explicitly transfer the distilled knowledge to $\mathcal{T}_i$.


An overview of our approach is presented in Figure~\ref{fig:figureoverview}, where WMB takes $\mathcal{T}_i$ and $\mathcal{S}_i$ as inputs that belong to $\mathbb{R}^{h\times w \times c}$. 
The correlation of $\mathcal{T}_i$ and $\mathcal{S}_i$ is utilized to generate the coarse localization mask, with the cosine similarity function being used for this purpose. Specifically, the similarity between $x$ and $y$ positions in the feature map is defined as
\begin{equation}
	\mathcal{W}_i^{x,y} = \frac{{\mathcal{T}_i^{x,y} \cdot \mathcal{S}_i^{x,y}}}{{\left\| {\mathcal{T}_i^{x,y}} \right\| \times \left\| {\mathcal{S}_i^{x,y}} \right\|}},
\end{equation}
where $\mathcal{T}_i^{x,y} \in \mathbb{R}^{1\times 1 \times c}$, $\mathcal{S}_i^{x,y} \in \mathbb{R}^{1\times 1 \times c}$. 
The similarity of each position is then evaluated to form the matrix $\mathcal{W}_i = \{\mathcal{W}_i^{x,y} |  x\in [1, h], y\in [1, w]\}$. Since the anomalous features in two networks are inclined to be unaligned, $1 - \mathcal{W}_i$  represents the coarse localization of the anomalous regions and servers as an abnormal coarse localization mask. 
Finally, utilizing explicit weighting, WMB transfers the coarse localization information related to anomalies obtained from distilled knowledge to $\mathcal{T}_i$, with this process being expressed as:
\begin{equation}
	{\mathcal{C}_i} = {(1-\mathcal{W}_i})\cdot {\mathcal{T}_i}, \label{eq3}
\end{equation}
where $\mathcal{C}_i$ denotes the outputs generated by WMB. The weighted approach is more suitable for the ADPS than the feature fusion approach, as described in the ablation experiment (\textit{Ref.} Sec \ref{sec4.D}). Therefore, $\mathcal{C}_i$ contains rich knowledge obtained through the pre-trained model's prior knowledge while simultaneously providing the ability to roughly discriminate between abnormal and normal regions based on distilled knowledge.

\subsection{Post-Segmentation Module}\label{sec3.4}

Traditional KDAD methods often infer the difference obtained from the distillation model as the result of anomaly segmentation. This results in two drawbacks, one is that the features extracted from the pre-trained network are only used as a ``reference standard" and not explored further for their potential, and another is that the segmentation results obtained yield low resolution with low accuracy.
To solve these drawbacks simply and effectively, we leverage the advantages of $\mathcal{C}_i$, which retains both prior and distilled knowledge, and develop the Post-Segmentation Module (PSM) to identify the normal and abnormal distribution from $\mathcal{C}_i$. PSM seeks to segment the abnormal regions with high accuracy and clear boundaries. 
Specifically, PSM employs the multiple UpBlock layers of UNet~\cite{DBLP:conf/miccai/RonnebergerFB15} as the backbone. To fully exploit the features extracted from each layer of the pre-trained teacher model, we use the weighted multiscale feature map $C=\{\mathcal{C}_1,\mathcal{C}_2,...,\mathcal{C}_n\}$ as the input. Here, $n$ represents the number of multiscale layers. We present UpBlock $i$ as an example to illustrate the upsampling process.

In UpBlock layer $i$ of PSM, the output $\mathcal{P}_{i-1}$ of UpBlock $i-1$ is fed as the input:
\begin{equation}
	{\mathcal{P}_i^m} = TrConv_i(\mathcal{P}_{i-1}),
\end{equation}
where $TrConv_i(\cdot)$ denotes the $2\times$ transposed convolution, and $\mathcal{P}_i^m$ is the output of the transposed convolution. Subsequently,  $\mathcal{P}_i^m$ and $\mathcal{C}_{i}$ are concatenated together along the channel dimension.
 \begin{equation}
 	{\mathcal{P}_i^c} = cat(\mathcal{P}_i^m,\mathcal{C}_{i}),
 \end{equation}
where $cat(\cdot)$ denotes the concatenation of two feature maps along the channel dimension.
Finally, we obtain the output $\mathcal{P}_i$ of UpBlock $i$ via the convolution layer.
 \begin{equation}
	{\mathcal{P}_{i+1}} = \sigma(BN_i^1(Conv_i^1(\sigma(BN_i^2(Conv_i^2(\mathcal{P}_i^c)))),
\end{equation}
where $Conv_i^1(\cdot)$ and $Conv_i^2(\cdot)$ indicates two $3 \times 3$ convolution layers with stride $1$ in UpBlock $i$ layer,
$BN_i^1(\cdot)$ and $BN_i^2(\cdot)$ denotes two batch normalization layers, $\sigma(\cdot)$ denotes the ReLU activation function. Finally, the output $\mathcal{P}_{n}$ is convolved by $3\times3$ convolution layer to obtain the segmentation mask $M$ with the size of $H \times W$.
 \begin{equation}
	{M} = Softmax(Conv_f(\mathcal{P}_{n})),
\end{equation}
where $Softmax(\cdot)$ denotes the Softmax function, and $Conv_f(\cdot)$ represents $3 \times 3$ convolution layer.

\subsection{Loss Function}\label{section3.5}

To train ADPS, this work follows the DRAEM \cite{zavrtanik2021draem} to generate simulated anomaly samples. Additionally, Distillation Loss and Focal Loss \cite{lin2017focal} are introduced as part of the loss function. 
The Distillation Loss, defined as follows, is used in the model training:
\begin{equation}
\resizebox{0.9\hsize}{!}{$
	\mathcal{L}_d =\sum\limits_{i = 1}^{n}{ (1 - Y) \cdot (1 - \frac{{{T_i} \cdot {S_i}}}{{\left\| {{\mathcal{T}_i}} \right\| \times \left\| {{\mathcal{S}_i}} \right\|}}) + Y \cdot \frac{{{\mathcal{T}_i} \cdot {\mathcal{S}_i}}}{{\left\| {{\mathcal{T}_i}} \right\| \times \left\| {{\mathcal{S}_i}} \right\|}}},
$}
\end{equation}
where $Y$ denotes the Ground Truth (GT) label of training samples such that $Y = \{ {y_{u,v}} \in \{ 0,1\} |u \in [1,h],v \in [1,w]\} $. Furthermore, the proposed approach utilizes the Focal Loss ${\cal L}_{s}$ \cite{lin2017focal} defined as follows:
\begin{equation}
	{\cal L}_{s} = \left\{ {\begin{array}{*{20}{c}}
			{ - {{(1 - p_{i,j})}^\tau }\log (p_{i,j}),}&{y_{i,j} = 1}\\
			{ - {p_{i,j}^\tau }\log (1 - p_{i,j}),}&{y_{i,j} = 0}
	\end{array}}, \right.\label{equ8}
\end{equation}
where $p_{i,j}$ denotes the probability of the abnormality in the coordinate $(i,j)$ of $M$. To incorporate the distillation loss ${\cal L}_{d}$ and segmentation loss ${\cal L}_{s}$ in the training process of ADPS, the overall loss $\mathcal{L}$ is defines as follows:

\begin{equation}
	\mathcal{L}=\mathcal{L}_d+\lambda\mathcal{L}_s,
\end{equation}
where $\lambda$ is set to control the importance of the two losses.

\begin{table*}[t!]
	\centering
	\renewcommand\arraystretch{1.5}
	\caption{
		The anomaly classification results in terms of $\mathcal{AUROC}_{cla}$ (\%) on the MVTec AD dataset. The best results are marked in bold.}
  \vspace{-2mm}
	\resizebox{\textwidth}{!}{%
		\setlength{\tabcolsep}{1mm}{
			\begin{tabular}{c|ccccc|c|cccccccccc|c|c}
				\toprule
				\multirow{2}{*}{Methods} & \multicolumn{5}{c}{Textures}                                                &                        & \multicolumn{10}{c}{Objects}                                                                                                                               &                        & \multirow{2}{*}{\textbf{MEAN}} \\ \cline{2-18}
				& Carpet        & Grid         & Leather      & Tile          & Wood          & \textit{\textbf{mean}} & Bottle       & Cable         & Capsule       & Hazelnut      & Metal nut     & Pill          & Screw         & Toothbrush   & Transistor    & Zipper       & \textit{\textbf{mean}} &                                \\ \hdashline
				U-Std   \cite{bergmann2020uninformed}                 & 91.6          & 81.0         & 88.2         & 99.1          & 97.7          & 91.5                   & 99.0         & 86.2          & 86.1          & 93.1          & 82            & 87.9          & 54.9          & 95.3         & 81.8          & 91.9         & 85.8                   & 87.7                           \\
				SPADE      \cite{cohen2020sub}              & -             & -            & -            & -             & -             & -                      & -            & -             & -             & -             & -             & -             & -             & -            & -             & -            & -                      & 85.5                           \\
				MKDAD     \cite{salehi2021multiresolution}               & 79.3          & 78           & 95.1         & 91.6          & 94.3          & 87.7                   & 99.4         & 89.2          & 80.5          & 98.4          & 73.6          & 82.7          & 83.3          & 92.2         & 85.6          & 93.2         & 87.8                   & 87.7                           \\
				DAAD+     \cite{hou2021divide}               & 86.6          & 95.7         & 86.2         & 88.2          & 98.2          & 91.0                   & 97.6         & 84.4          & 76.7          & 92.1          & 75.8          & 90            & 98.7          & 99.2         & 87.6          & 85.9         & 88.8                   & 89.5                           \\
				RIAD     \cite{zavrtanik2021reconstruction}               & 84.2          & 99.6         & \textbf{100.0} & 98.7          & 93            & 95.1                   & 99.9         & 81.9          & 88.4          & 83.3          & 88.5          & 83.8          & 84.5          & \textbf{100.0} & 90.9          & 98.1         & 89.9                   & 91.7                           \\
				MAD    \cite{rippel2021modeling}                  & 95.5          & 92.9         & \textbf{100.0} & 97.4          & 97.6          & 96.7                   & \textbf{100.0} & 94.0          & 92.3          & 98.7          & 93.1          & 83.4          & 81.2          & 95.8         & 95.9          & 97.9         & 93.2                   & 94.4                           \\
				MF \cite{wu2021learning}&94.0 &85.9&99.2&99.0&\textbf{99.2}&95.5&99.1&97.1&87.5&99.4&96.2&90.1&\textbf{97.5}&\textbf{100.0}&94.4&98.6&96.0&95.8\\
				Cutpaste     \cite{li2021cutpaste}            & 93.1          & 99.9         & \textbf{100.0} & 93.4          & 98.6          & 97.0                   & 98.3         & 80.6          & \textbf{96.2} & 97.3          & 99.3          & 92.4          & 86.3          & 98.3         & 95.5          & 99.4         & 94.4                   & 95.2                           \\
				PaDim      \cite{defard2021padim}              & \textbf{99.8} & 96.7         & \textbf{100.0} & 98.1          & \textbf{99.2} & 98.8                   & 99.9         & 92.7          & 91.3          & 92.0          & 98.7          & 93.3          & 85.8          & 96.1         & 97.4          & 90.3         & 93.8                   & 95.5                           \\
				AnoSeg      \cite{DBLP:journals/corr/abs-2110-03396}             & 96.0          & 99.0         & 99.0         & 98.0          & 99.0          & 98.2                   & 98.0         & \textbf{98.0} & 84.0          & 98.0          & 95.0          & 87.0          & 97.0 & 99.0         & 96.0          & 99.0         & 95.1                   & 96.0                           \\ \hdashline
				\textbf{ADPS}            & 97.4          & \textbf{100.0} & \textbf{100.0} & \textbf{99.8} & 97.7          & \textbf{99.0}          & \textbf{100.0} & 93.7          & 96.0          & \textbf{99.6} & \textbf{99.7} & \textbf{95.3} & 89.5          & 95.3         & \textbf{97.6} & \textbf{100.0} & \textbf{96.7}          & \textbf{97.4}                  \\ \bottomrule
			\end{tabular} 
	}}\label{TAB.1} 
 \vspace{-2mm}
\end{table*}

\section{Experiments}~\label{section4}
\subsection{Experimental Settings}

\subsubsection{Datasets}
To demonstrate the effectiveness of our proposed ADPS, we conducted experiments
on three datasets: MVTec AD~\cite{bergmann2019mvtec}, KolektorSDD~\cite{tabernik2020segmentation}, and KolektorSDD2~\cite{bovzivc2021mixed}.
(1) The MVTec AD~\cite{bergmann2019mvtec} dataset consists of $15$ classes of high-resolution images derived from actual industrial scenarios, subdivided into five categories of texture images (referred to as "Texture") and ten categories of object images (referred to as "Object"). The training dataset contains only normal images, while the testing dataset includes a variety of anomaly patterns, such as scratches, defects, etc. This dataset presents a considerable challenge due to the subtle anomalies and high-resolution nature. 
(2) The KolektorSDD~\cite{tabernik2020segmentation} dataset consists of $399$ images, each with dimensions of $500 \times 1250$. These images were acquired under controlled conditions in a real-world industrial setting. 
(3) The KolektorSDD2~\cite{bovzivc2021mixed} is a comprehensive surface defect detection dataset containing over 3000 images, with each having a size of approximately $230 \times 630$. This dataset exposes a range of defect types commonly found in the industry, including scratches, dots, and surface defects.

\subsubsection{Implementation Details} 
The input image resolution is fixed at $256 \times 256$, while $k$ is set to $8$. In the training phase, the learning rate is set to $0.0001$ and the batch size is set to $32$. The network is trained for 300 epochs by the Adam optimizer~\cite{DBLP:journals/corr/KingmaB14}, with a learning rate decay at epochs 240 and 270 with a decay rate of $0.2$. We set the value of $\lambda$ to $1$.

\subsubsection{Baseline Approaches}
We compare our ADPS against several baselines, including: U-Std~\cite{bergmann2020uninformed}, MKDAD~\cite{salehi2021multiresolution},MF~\cite{wu2021learning}, RD~\cite{deng2022anomaly}, DAAD+~\cite{hou2021divide}, RIAD~\cite{zavrtanik2021reconstruction}, Cutpaste~\cite{li2021cutpaste}, DRAEM~\cite{zavrtanik2021draem}, DRAEM+SSPCAB~\cite{DBLP:conf/cvpr/RisteaMINKMS22}, AnoSeg~\cite{DBLP:journals/corr/abs-2110-03396}, SGSF~\cite{xing2022self}, SPADE~\cite{cohen2020sub}, PaDim~\cite{defard2021padim}, MAD~\cite{rippel2021modeling}, and Semi-orthogonal~\cite{kim2021semi}. Among these, SPADE, PaDim, Semi-orthogonal, and RD utilize WideResNet50~\cite{DBLP:conf/bmvc/ZagoruykoK16} as the backbone, while MF introduces a Transformer-based structure. DRAEM, DRAEM+SSPCAB, Cutpaste, and our ADPS introduce simulated anomalous samples.


\subsubsection{Evaluation Metrics}
 As per previous studies~\cite{defard2021padim,deng2022anomaly,tao2022deep}, commonly used evaluation metrics for anomaly classification tasks is Area Under the Receiver Operating Characteristic (AUROC), and for anomaly segmentation tasks are Per-Region-Overlap (PRO) and AUROC. 
 However, these metrics inadequately reflect the actual segmentation performance when dealing with small anomalous regions due to the overwhelming number of normal pixels. 
 Towards this end,  Tao~\etal~\cite{tao2022deep} introduces a more convincing metric, Average Precision (AP), to reflect the performance of the proposed method for pixel-level segmentation. 
In this work, the performance of anomaly classification is evaluated using $\mathcal{AUROC}_{cla}$, while anomaly segmentation performance is measured using $\mathcal {AUROC}_{seg}$, $\mathcal{PRO}_{seg}$, and $ \mathcal {AP}_{seg}$ collectively. Notable, on account of the significantly lower number of abnormal pixels than normal ones, $ \mathcal {AP}_{seg}$ better indicates the segmentation accuracy.


\subsection{Comparison with State-of-the-art Methods}
\subsubsection{MVTec AD}
Table~\ref{TAB.1} illustrates the anomaly classification results obtained on the MVTec AD dataset. ADPS achieves very competitive anomaly classification results, with $100\%$ classification accuracy achieved across four categories. Significantly, ADPS shows notable performance improvement in several categories, outperforming the methods employing WideResNet-50 with improved accuracy by approximately $3-10\%$. These results demonstrate that ADPS makes full use of the pre-trained model's knowledge to achieve efficient anomaly classification. In comparison to Cutpaste~\cite{li2021cutpaste}, which further introduces simulated anomalous samples, ADPS obtains a superior performance gain of $2.7\%$. 
Thus, the strategy of leveraging asymmetric distillation to achieve coarse localization and fine segmentation of anomalies proves more effective than training the classifier directly. Overall, the experimental results highlight that ADPS, integrating distilled and prior knowledge, provides a more accurate normal image distribution and delivers powerful anomaly classification capability.




\begin{table*}[]
	\centering
	\renewcommand\arraystretch{1.5}
	\caption{
		The anomaly segmentation results in terms of $\mathcal{AUROC}_{seg}$, $\mathcal{PRO}_{seg}$, and $\mathcal{AP}_{seg}$ (\%) on the MVTec AD dataset. The proposed ADPS outperforms recent methods by nearly 10\% in terms of $\mathcal{AP}_{seg}$. The best results are marked in bold.}
  \vspace{-2mm}
	\resizebox{\textwidth}{!}{%
		\setlength{\tabcolsep}{0.8mm}{
			\resizebox{\columnwidth}{!}{%
				\begin{tabular}{l|c|ccccc|c|cccccccccc|c|c}
					\hline
					\multicolumn{1}{c|}{\multirow{2}{*}{Metric}}  & \multirow{2}{*}{Methods} & \multicolumn{5}{c}{Textures}                                                  &                        & \multicolumn{10}{c}{Objects}                                                                                                                                  &                        & \multirow{2}{*}{\textbf{MEAN}} \\ \cline{3-19}
					\multicolumn{1}{c|}{}                         &                          & Carpet        & Grid          & Leather       & Wood          & Tile          & \textit{\textbf{mean}} & Bottle        & Cable         & Capsule       & Hazelnut      & Metal nut     & Pill          & Screw         & Toothbrush    & Transistor    & Zipper        & \textit{\textbf{mean}} &                                \\ \hdashline
					\multirow{11}{*}{$\mathcal{AUROC}_{seg}$}                      & U-Std ~\cite{bergmann2020uninformed}                  & 93.5          & 89.9          & 97.8          & 92.1          & 92.5          & 93.2                   & 97.8          & 91.9          & 96.8          & 98.2          & 97.2          & 96.5          & 97.4          & 97.9          & 73.7          & 95.6          & 94.3                   & 93.9                           \\
					& RIAD    ~\cite{zavrtanik2021reconstruction}                & 96.3          & 98.8          & 99.4          & 85.8          & 89.1          & 93.9                   & 98.4          & 84.2          & 92.8          & 96.1          & 92.5          & 95.7          & 98.8          & 98.9          & 87.7          & 97.8          & 94.3                   & 94.2                           \\
					& Cutpaste   ~\cite{li2021cutpaste}          & 98.3          & 97.5          & 99.5          & 95.5          & 90.5          & 96.3                   & 97.6          & 90            & 97.4          & 97.3          & 93.1          & 95.7          & 96.7          & 98.1          & 93            & 99.3          & 95.8                   & 96.0                           \\
					& SPADE   ~\cite{cohen2020sub}                & 97.5          & 93.7          & 97.6          & 88.5          & 87.4          & 92.9                   & 98.4          & 97.2          & \textbf{99.0} & 99.1          & 98.1          & 96.5          & 98.9          & 97.9          & 94.1          & 96.5          & 97.6                   & 96.5                           \\
					& PaDim    ~\cite{defard2021padim}              & 99.0          & 98.6          & 99            & 94.1          & 94.1          & 97.0                   & 98.2          & 96.7          & 98.6          & 98.1          & 97.3          & 95.7          & 94.4          & 98.8          & 97.6          & 98.4          & 97.4                   & 97.4                           \\
					& DRAEM     ~\cite{zavrtanik2021draem}              & 95.5          & 99.7          & 98.6          & 96.4          & 99.2          & 97.9                   & 99.1          & 94.7          & 94.3          & 99.7          & \textbf{99.5} & 97.6          & 97.6          & 98.1          & 90.9          & 98.8          & 97.0                   & 97.3                           \\
					& SGSF~\cite{xing2022self}                    & 97.4          & 99.2          & 98.9          & 97.3          & 99.7          & 98.5                   & 96.3          & 93.3          & 97.4          & 99.5          & 99.5          & 99.5          & 97.2          & \textbf{99.6}          & 84.7          & 98.2          & 96.5                   & 97.2                           \\
					& MKDAD   ~\cite{salehi2021multiresolution}              & 95.6          & 91.8          & 98.1          & 82.8          & 84.8          & 90.6                   & 96.3          & 82.4          & 95.9          & 94.6          & 86.4          & 89.6          & 96.0          & 96.1          & 76.5          & 93.9          & 90.8                   & 90.7                           \\
					& RD     ~\cite{deng2022anomaly}                  & 98.9          & \textbf{99.3} & 99.4          & 95.6          & 95.3          & 97.7                   & 98.7          & \textbf{97.4} & 98.7          & 98.9          & 97.3          & 98.2          & \textbf{99.3} & 99.1 & \textbf{92.5} & 98.2          & \textbf{97.8}          & 97.8                           \\ \cline{2-20}
					& \textbf{ADPS}                     & \textbf{99.5} & 99.2          & \textbf{99.9} & \textbf{99.3} & \textbf{99.6} & \textbf{99.5}          & \textbf{99.5} & 94.6          & 94.1          & \textbf{99.6} & 97.5          & \textbf{99.3} & 98.7          & 99.1 & 92.2          & \textbf{99.6} & 97.4                   & \textbf{98.1}                  \\ \hdashline

\multirow{6}{*}{$\mathcal{PRO}_{seg}$}& U-Std~\cite{bergmann2020uninformed}   & 87.9 & 95.2 & 94.5 & 94.6 & 91.12 & 92.7 & 93.1 & 81.8 & 96.8 & \textbf{96.5} & \textbf{94.5} & 96.1 & 94.2 & 93.3 & 66.6 & 95.1 & 90.8 & 91.4 \\
 & MF~\cite{wu2021learning}  & 87.8 & 86.5 & 95.9 & 88.1 & 84.8 & 88.6 & 88.8 & \textbf{93.7} & 87.9 & 88.6 & 86.9 & 93.0 & 95.4 & 87.7 & \textbf{92.6} & 93.6 & 90.8 & 90.1 \\
 & SPADE~\cite{cohen2020sub}   & 94.7 & 86.7 & 97.2 & 75.9 & 75.9 & 87.4 & 95.5 & 90.9 & 93.7 & 95.4 & 94.4 & 94.6 & 96.0 & 93.5 & 87.4 & 92.6 & \textbf{93.4} & 91.7 \\
 & PaDim~\cite{defard2021padim} & 96.2 & 94.6 & 97.8 & 86.0 & 91.1 & 93.2 & 94.8 & 88.8 & 93.5 & 92.6 & 85.6 & 92.7 & 94.4 & 93.1 & 84.5 & 95.9 & 91.6 & 92.1 \\
 & RD~\cite{deng2022anomaly}  & 97.0 & \textbf{97.6} & 99.1 & 90.6 & 90.9 & 95.0 & 96.6 & 91.0 & \textbf{95.8} & 95.5 & 92.3 & \textbf{96.4} & \textbf{98.2} & \textbf{94.5} & 78.0 & 95.4 & \textbf{93.4} & 93.9 \\ \cline{2-20}
 & \textbf{ADPS} & \textbf{97.5} & 97.2 & \textbf{99.3} & \textbf{96.6} & \textbf{97.9} & \textbf{97.7} & \textbf{97.8} & 82.9 & 93.4 & 95.2 & 91.5 & 96.0 & 94.3 & 93.4 & 85.4 & \textbf{98.3} & 92.8 & \textbf{94.4} \\ \hdashline

					\multirow{9}{*}{$\mathcal{AP}_{seg}$} & U-Std ~\cite{bergmann2020uninformed}                    & 52.2          & 10.1          & 40.9          & 53.3          & 65.3          & 44.4                   & 74.2          & 48.2          & 25.9          & 57.8          & 83.5          & 62.0            & 7.8           & 37.7          & 27.1          & 36.1          & 46.0                   & 45.5                           \\
					& RIAD    ~\cite{zavrtanik2021reconstruction}                  & 61.4          & 36.4          & 49.1          & 38.2          & 52.6          & 47.5                   & 76.4          & 24.4          & 38.2          & 33.8          & 64.3          & 51.6          & 43.9          & 50.6          & 39.2          & 63.4          & 48.6                   & 48.2                           \\
					& PaDim     ~\cite{defard2021padim}              & 60.7          & 35.7         & 53.5          & 46.3          & 52.4          & 49.7                   & 77.3          & 45.4          & 46.7          & 61.1          & 77.4          & 61.2          & 21.7          & 54.7          & 72.0            & 58.2          & 57.6                   & 55.0                             \\
					& DRAEM      ~\cite{zavrtanik2021draem}              & 53.5          & \textbf{65.7}         & 75.3          & 77.7          & 92.3          & 72.9                   & 86.5          & 52.4          & 49.4          & \textbf{92.9} & 96.3 & 48.5          & 58.2          & 44.7          & 50.7          & 81.5          & 66.1                   & 68.4                           \\
& DRAEM +SSPCAB\cite{DBLP:conf/cvpr/RisteaMINKMS22} & 59.4 & 61.1 & 76.0 & 95.0 & 77.1 &  73.7& 87.9 & 57.2 & 50.2 & 92.6 & \textbf{98.1} & 52.4 & \textbf{72.0} & 51.0 & 48.0 & 77.1 & 68.7 & 69.9 \\
     & RD~\cite{deng2022anomaly}                       & 64.1          & 47.6          & 52.4          & 50.4          & 53.6          & 53.6                   & 79.4          & 59.2          & 45.8          & 64.5          & 80.9          & 80.0          & 54.8          & 54.5          & \textbf{55.7} & 60.6          & 63.5                   & 60.2                           \\ \cline{2-20} 
					& \textbf{ADPS}                     & \textbf{83.9} & 62.3          & \textbf{81.6} & \textbf{89.4} & \textbf{96.3}          & \textbf{82.7}          & \textbf{92.9} & \textbf{65.4} & \textbf{62.5} & 89.4          & 82.9          & \textbf{88.7}          & 59.1 & \textbf{67.5}          & 53.1          & \textbf{85.4} & \textbf{74.7}          & \textbf{77.4}                  \\ \cline{1-20} 
				\end{tabular}
	}}}\label{TAB.2} 
\vspace{-1mm}
\end{table*}

\begin{figure*}
	\centering
	\includegraphics[width=0.99\linewidth]{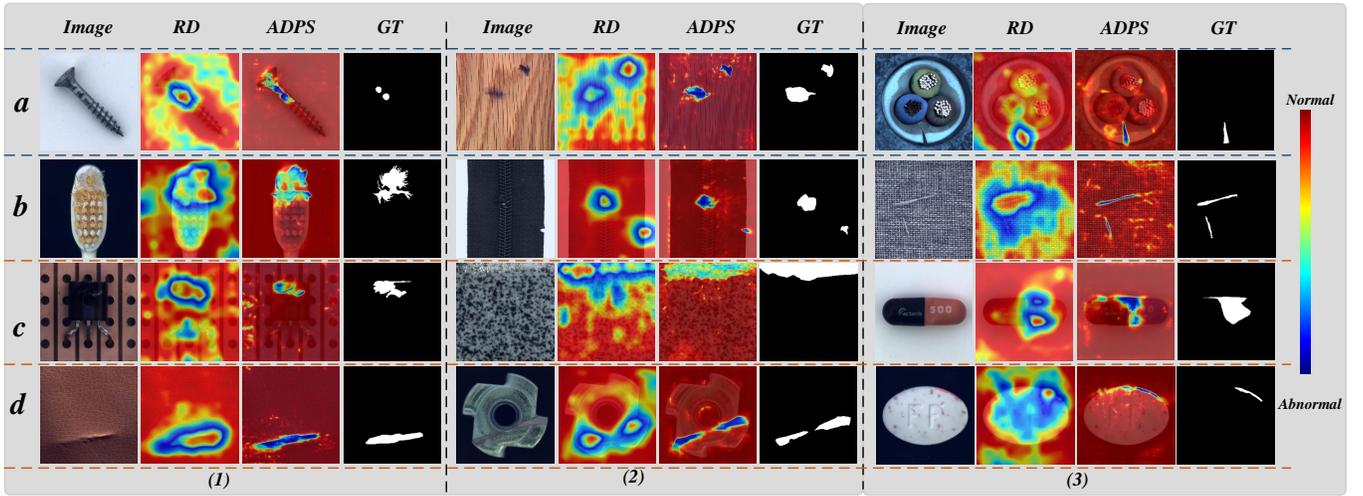}
 \vspace{-2mm}
	\caption{The anomaly segmentation results on the MVTec AD dataset. Compared with RD~\cite{deng2022anomaly}, ADPS achieves high-confidence segmentation results in both anomalous and normal regions with different categories.}
	\label{fig:figure41}
 \vspace{-2mm}
\end{figure*}

\begin{table}[t!]
	\centering
	\renewcommand\arraystretch{1.3}
	\caption{Anomaly detection results for the KolektorSDD and KolektorSDD2 datasets. }\vspace{-2mm}
		\resizebox{\columnwidth}{!}{%
			\setlength{\tabcolsep}{0.8mm}{
				\begin{tabular}{c|c|c|c|c}
					\hline
					\multirow{2}{*}{Method} & KolektorSDD       & \multicolumn{3}{c}{KolektorSDD2}      \\ \cline{2-5} 
					& $\mathcal{AUROC}_{seg}$ & $\mathcal{AUROC}_{cla}$ & $\mathcal{AUROC}_{seg}$ & $\mathcal{AP}_{seg}$ \\ \hline
					Semi-orthogonal  ~\cite{kim2021semi}                              & 96.0         & -          & 98.1     & -     \\
					PaDim ~\cite{defard2021padim}                                          & 94.5       & -          & 95.6     & -     \\
					U-Std   ~\cite{bergmann2020uninformed}                                       & 89.6       & -          & 95.0     & -     \\
     DRAEM~\cite{zavrtanik2021draem}                                      &-                                                                                     & 93.1                           & 93.4                           & 50.5                           \\
					SGSF      ~\cite{xing2022self}                                     & -          & 93.5       & 91.5     & 51.6  \\
					RD   ~\cite{deng2022anomaly}  
					& -          & 94.8       & 98.2     & 47.7 \\ \hline
					\textbf{ADPS}                                            & \textbf{96.3}       & \textbf{95.6}       & \textbf{99.2}     & \textbf{72.5}  \\ \hline
				\end{tabular}
	}}\label{TAB.4}
 \vspace{-2mm}
\end{table}

In addition to anomaly classification, we demonstrate the anomaly segmentation performance of ADPS  and compare it to current state-of-the-art methods. The evaluation is based on three key metrics on the MVTec AD dataset, with the results presented in Table~\ref{TAB.2}. Our approach, ADPS, outperforms knowledge distillation-based methods such as U-Std and MKDAD by $5-10\%$ on $\mathcal{AUROC}_{seg}$ and $3\%$ on $\mathcal{PRO}_{seg}$. It also exceeds U-Std by $32\%$ on $\mathcal{AP}_{seg}$, demonstrating its excellent anomaly segmentation capability. Compared to recent distillation methods, the superior performance of ADPS is attributed to its asymmetric distillation paradigm, which shows effectiveness in handling anomalous images. 
The representations extracted from the student network of input local images are harder to match with those of the teacher network of input global images. This leads to better detection of intricate anomalous regions, resulting in a significant improvement in segmentation precision. 
ADPS also surpasses DRAEM, DRAEM+SSPCAB, and Cutpaste methods that rely on forged anomalous samples, even though DRAEM (including combined SSPCAB) have larger parameters of networks, achieving an impressive $\mathcal{AP}_{seg}$ of $77.4\%$. In particular, DRAEM similarly uses a segmentation module, while ADPS significantly outperforms its anomaly segmentation capability. The competitive results of ADPS are attributed to its integration of two fundamental components: harnessing the feature representation capabilities inherent in the teacher network and using distilled knowledge to obtain a coarse localization of anomalous regions followed by precise anomaly segmentation.

Figure~\ref{fig:figure41} shows the visualization image of anomaly segmentation on the MVTec AD dataset. Subfigures $a$-\textit{(1)}, $a$-\textit{(2)}, and $a$-\textit{(3)} demonstrate the anomalous region segmentation capability of RD and ADPS. ADPS utilizes the segmentation module to improve its segmentation accuracy by exploiting coarse localization features from the distilled knowledge. Conversely, RD approximates the location of anomalous regions but is accompanied by significant noise. Subfigures $c$-\textit{(1)}, $d$-\textit{(2)}, and $d$-\textit{(3)} demonstrate the high resemblance between anomalous and normal regions, which poses challenges to the efficient segmentation of current anomaly detection methods. However, the exceptional ability of ADPS in anomaly segmentation stands out.


\subsubsection{KolektorSDD\&KolektorSDD2}
\begin{figure}
	\centering
	\includegraphics[width=0.99\linewidth]{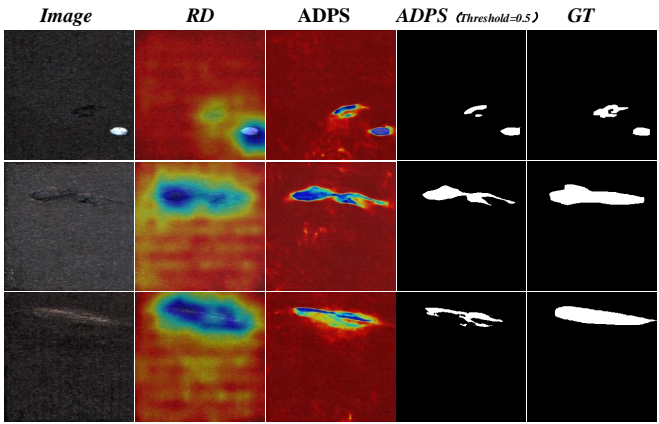}
	\caption{The anomaly segmentation results on the KolektorSDD2 dataset.}
	\label{fig:figure5} 
 \vspace{-2mm}
\end{figure}

To further demonstrate the effectiveness of ADPS, we conduct anomaly classification and anomaly segmentation experiments on two additional benchmark datasets (\ie, KolektorSDD and KolektorSDD2). The results, presented in Table~\ref{TAB.4}, show that ADPS achieves state-of-the-art performance on both datasets. 
Notably, for the KolektorSDD2 dataset, ADPS achieves an $\mathcal{AP}_{seg}$ of $72.5\%$, which is a substantial improvement of $20\%$ and $21\%$ over the performance of DRAEM and SGSF, respectively. It is worth mentioning that both DRAEM and SGSF employ forged anomaly samples and segmentation frameworks. This significant advantage in performance can be attributed to the thorough exploration of the potential of pre-trained models by ADPS for anomaly detection tasks. 
In order to achieve accurate anomaly segmentation, distilled knowledge, and prior knowledge are utilized to generate coarse localization information, which is further utilized by the segmentation module. 
The results of anomaly segmentation are shown in Figure~\ref{fig:figure5}. Exemplary segmentation of anomaly contours and normal regions, along with high-confidence segmentation results, is achieved by ADPS, outperforming RD in this regard. Furthermore, by uniformly calibrating the segmentation threshold at $0.5$ across all categories, ADPS obtains precise anomaly segmentation maps.



\begin{table*}[t!]
	\centering
	\renewcommand\arraystretch{1.3}
	\caption{The performance of four structures on the MVTec AD dataset. ADPS is the best in terms of anomaly classification and anomaly segmentation performance. }\vspace{-2mm}
			\setlength{\tabcolsep}{2mm}
\begin{tabular}{c|cccc|cccc|cccc}
					\hline
					\multicolumn{1}{c|}{\multirow{2}{*}{Category}} & \multicolumn{4}{c|}{$\mathcal{AP}_{seg}$}                     & \multicolumn{4}{c|}{$\mathcal{AUROC}_{cla}$}                   & \multicolumn{4}{c}{$\mathcal{AUROC}_{seg}$}                     \\ \cline{2-13}
					\multicolumn{1}{c|}{} 
					& W/O.PW & W/O.T & W/O.S & ADPS                                                & W/O.PW                          & W/O.T & W/O.S                           & ADPS                         & W/O.PW   & W/O.T                                             & W/O.S                          & ADPS                           
					          \\ \hline
					\multicolumn{1}{c|}{Capsule}    & 28.9   & 27.4   & 56.1  & \multicolumn{1}{c|}{\textbf{62.6}} & 70.5                            & 88.2   & \textbf{97.2}  & 96.3          & \textbf{97.7}                  & 88.0  & 93.0                           & 97.2                           \\
					\multicolumn{1}{c|}{Bottle}     & 82.8   & 60.0   & 89.7  & \multicolumn{1}{c|}{\textbf{93.9}} & \textbf{100.0} & 91.8   & \textbf{100.0} & \textbf{100.0}  & 98.7       & 93.6                    & 99.0                           & \textbf{99.5} \\
					\multicolumn{1}{c|}{Carpet}     & 66.4   & 73.6   & \textbf{84.3}  & \multicolumn{1}{c|}{83.7} & 95.7                            & 100.0  & 99.6                            & \textbf{97.0}   & 99.0      & 98.8                     & \textbf{99.6} & 99.5                           \\
					\multicolumn{1}{c|}{Leather}    & 68.9   & \textbf{81.4}   & 80.6  & \multicolumn{1}{c|}{80.4} & 99.8                            & 100.0  & \textbf{100.0} & \textbf{100.0}  & 99.7      & 99.9                     & \textbf{99.9} & 99.8                           \\
					\multicolumn{1}{c|}{Pill}       & 65.2   & 78.3   & \textbf{89.6}  & \multicolumn{1}{c|}{85.9} & 77.6                            & 89.8   & \textbf{94.8}  & 94.7                             & 96.7     & 98.3                      & \textbf{99.3} & 99.0                           \\
					\multicolumn{1}{c|}{Transistor} & 29.3   & 30.1   & 28.9  & \multicolumn{1}{c|}{\textbf{48.3}} & 85.4                            & 82.5   & 89.8                            & 95.5                             & 80.6  & 74.2                         & 66.3                           & \textbf{84.8} \\
					\multicolumn{1}{c|}{Tile}       & 82.1   & 95.2   & \textbf{96.5}  & \multicolumn{1}{c|}{96.1} & 98.6                            & 99.6   & \textbf{100.0} & 99.7                             & 97.5     & 99.5                      & \textbf{99.6} & 99.5                           \\
					\multicolumn{1}{c|}{Cable}      & 49.7   & 43.5   & 51.3  & \multicolumn{1}{c|}{\textbf{63.3}} & 86.6                            & 78.9   & 91.2                            & \textbf{94.9}   & \textbf{95.6} & 86.2 & 89.7                           & 94.1                           \\
					\multicolumn{1}{c|}{Zipper}     & 57.7   & 66.4   & 85.6  & \multicolumn{1}{c|}{\textbf{86.0}} & 94.5                            & 99.9   & \textbf{100.0} & \textbf{100.0}  & 98.2      & 98.8                     & \textbf{99.6} & 99.5                           \\
					\multicolumn{1}{c|}{Toothbursh} & 40.6   & 47.2   & \textbf{67.0}  & \multicolumn{1}{c|}{65.7} & 71.1                            & 86.1   & 98.1                            & \textbf{99.4}   & 98.0      & 96.7                     & \textbf{99.2} & \textbf{99.2} \\
					\multicolumn{1}{c|}{Metal\_nut} & 82.1   & 90.4   & 58.3  & \multicolumn{1}{c|}{\textbf{87.1}} & 99.6                            & 99.2   & 99.6                            & \textbf{100.0} & \textbf{98.2} & 98.4 & 89.3                           & 98.1                           \\
					\multicolumn{1}{c|}{Hazelnut}   & 77.9   & 78.8   & 86.5  & \multicolumn{1}{c|}{\textbf{90.5}} & \textbf{99.9}  & 93.6   & 99.7                            & 99.1                             & 99.2      & 99.1                     & 99.5                           & \textbf{99.6} \\
					\multicolumn{1}{c|}{Screw}      & 22.5   & 21.8   & 52.8  & \multicolumn{1}{c|}{\textbf{59.2}} & 84.9                            & 79.5   & 91.6                            & \textbf{94.2}   & 98.5        & 96.5                   & 97.4                           & \textbf{98.8} \\
					\multicolumn{1}{c|}{Grid}       & 36.7   & 56.6   & 62.9  & \multicolumn{1}{c|}{\textbf{65.6}} & 94.6                            & 99.2   & 99.8                            & \textbf{100.0} & 97.9        & 99.3                    & 99.3                           & \textbf{99.5} \\
					\multicolumn{1}{c|}{Wood}       & 77.9   & 78.4   & 88.4  & \multicolumn{1}{c|}{\textbf{89.8}} & \textbf{99.0}  & 97.9   & 98.0                            & 94.6                             & 97.9      & 96.8                     & 98.5                           & \textbf{99.0} \\ \hline
					\textbf{MEAN}  & 57.9   & 61.9   & 71.9  & \textbf{77.2}                      & 90.5                            & 92.4   & 97.3                            & \textbf{97.7}                       & 96.9    & 94.9                        & 95.3                           & \textbf{97.8} \\ \hline
				\end{tabular}
    \label{fig:graph8} 
 \vspace{-2mm}
\end{table*}


\subsection{Effectiveness of Asymmetric Distillation Paradigm} 

\subsubsection{Asymmetric Inputs}
\begin{figure}
	\centering
	\includegraphics[width=0.8\linewidth]{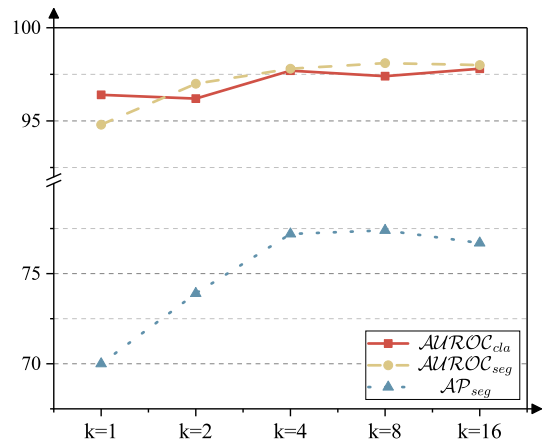}
  \vspace{-2mm}
	\caption{The performance of asymmetric distillation paradigm ($k=2,4,8,16$) and symmetric distillation paradigm ($k=1$).}
	\label{fig:graph6} 
 \vspace{-4mm}
\end{figure}

\begin{figure}
	\centering
	\includegraphics[width=0.9\linewidth]{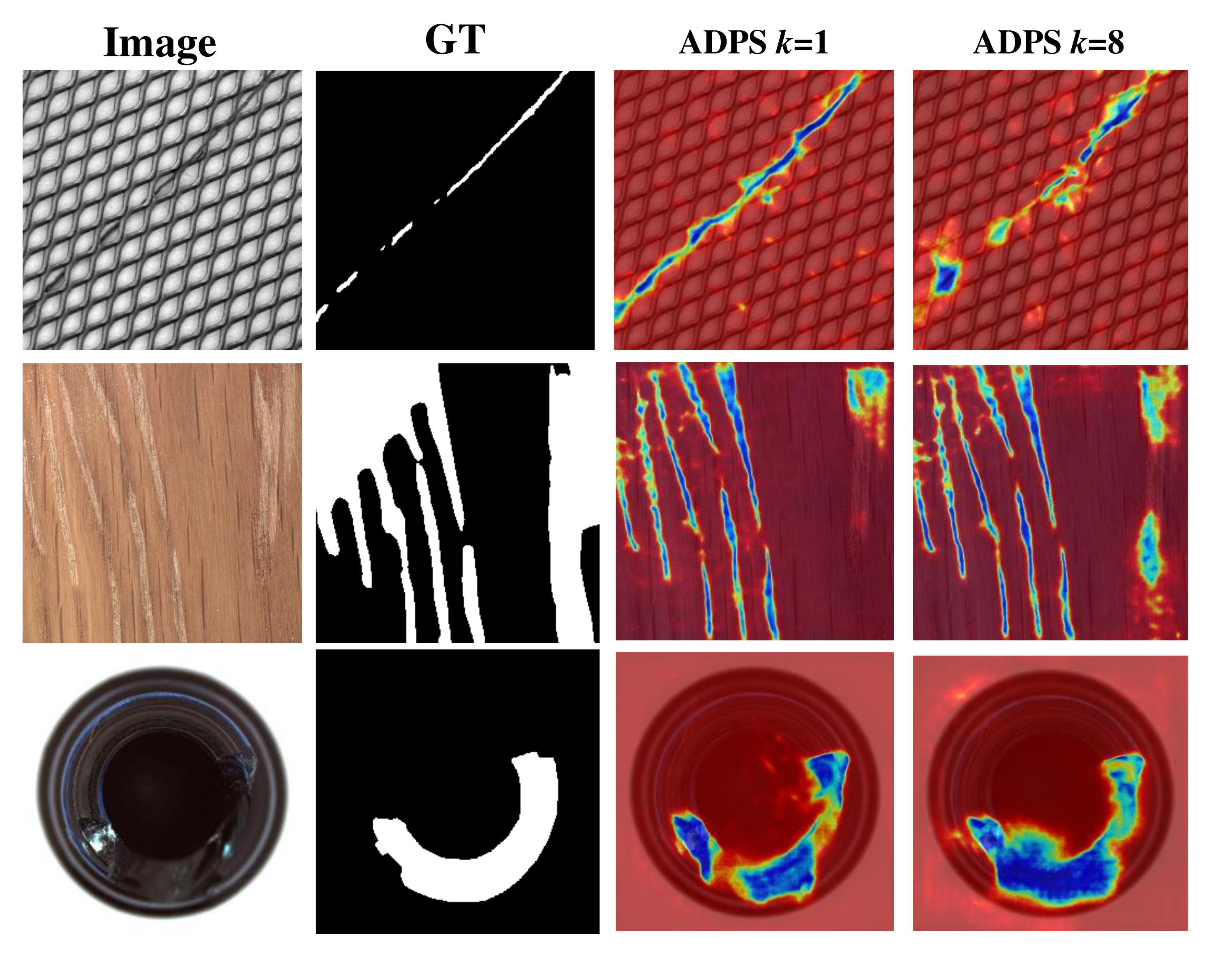}\vspace{-2mm}
	\caption{Visualization images using symmetric distillation ($k=1$) and asymmetric distillation methods ($k=8$). }
	\label{fig:figutre11} \vspace{-2mm}
\end{figure}

We first conduct experiments on the MVTec AD dataset to validate the influence of various inputs on the asymmetric distillation paradigm. When $k$=$1$, ADPS adopts the symmetric distillation paradigm. In the cases where $k$ belongs to $\{2,4,8,16\}$, ADPS utilizes the asymmetric distillation paradigm by dividing the input into $k \times k$ patches at the input layer, respectively. 
The experimental results are illustrated in Figure~\ref{fig:graph6}. 
The supremacy of the asymmetric distillation paradigm is visibly established, particularly in terms of $\mathcal{AP}_{seg}$. Moreover, the optimal performance is achieved when $k$=$4$ or $8$. The division into patches facilitates the focused attention of the student network on specific regions, enabling ADPS to recognize anomalous pixels that closely resemble normal ones. However, excessive division (when $k$=$16$) can cause a loss of crucial original structural information leading to a decrease in performance. We visualize the anomaly segmentation results for the symmetric distillation paradigm and the asymmetric distillation paradigm shown in Figure~\ref{fig:figutre11}. It is observed that the student network with an asymmetric input can better capture the anomalous features of tiny regions with detailed boundaries. In scenarios like the `Grid' category, the sequential learning of local images allows the student network to concentrate more intensely on local details, thereby facilitating a more accurate localization map for ``foreign objects”.



\subsubsection{Asymmetric Layers}
In addition to introducing the asymmetric inputs, ADPS introduces the asymmetric distillation paradigm at various feature levels in different stages. Taking WideResNet-50 serves as the backbone of the teacher network as an example, we introduce the asymmetric distillation strategy in Stage~1 and Stage~2, building upon the asymmetric input layer. The anomaly detection results are shown in Table~\ref{tav5-1}. 
The introduction of asymmetric distillation at the shallow layer (stage~1) improves the anomaly detection performance for texture images, although its effect on the ``Object" type is not substantial. 
However, the performance drops significantly upon further adoption at the deep layers (stage~2). This may occur due to the low resolution of segmented features in the deeper layers, which disrupts the original nearest-neighbor relationships within the image. Consequently, the destruction of structural information impairs the effectiveness of distillation in providing coarse localization information, thereby influencing the detection capability of ADPS.

\begin{table*}
	\centering
	\renewcommand\arraystretch{1.5}
	\caption{The results of introducing asymmetric knowledge paradigm for different stages with $k=4$. ``Stage i" indicates that at stage $i$. }
\vspace{-2mm}
				\begin{tabular}{ccc|ccc|ccc|ccc|ccc}				
					\toprule
					\multirow{2}{*}{Stage 0}& \multirow{2}{*}{Stage 1}& \multirow{2}{*}{Stage 2}& \multicolumn{3}{c|}{$\mathcal{AUROC}_{cla}$}                     & \multicolumn{3}{c|}{$\mathcal{AUROC}_{seg}$} &\multicolumn{3}{c|}{$\mathcal{PRO}_{seg}$}                  & \multicolumn{3}{c}{$\mathcal{AP}_{seg}$}                     \\ \cline{4-15}
					 &&&Texture             & Object    & Mean   & Texture             & Object    & Mean          & Texture             & Object    & Mean             & Texture             & Object    & Mean             \\ \hline
					\ding{52}&\ding{55} &\ding{55} & 98.2          & \textbf{97.4} & 97.7          & \textbf{99.5} & \textbf{97.0} & \textbf{97.8}   &97.8&90.6&93.0           & 83.1          & \textbf{74.2} & \textbf{77.2} \\
					\ding{52}&\ding{52} &\ding{55} & \textbf{99.3} & \textbf{97.4} & \textbf{98.0} & \textbf{99.5} & 96.7          & 97.6   &\textbf{98.0}&\textbf{91.0}&\textbf{93.3}       & \textbf{83.7} & 73.5          & 76.9          \\
					\ding{52}&\ding{52} &\ding{52} & 99.1          & 97.3 & 97.8          & 99.4 & 96.4          & 97.4    &\textbf{98.0}&89.9&92.6      & 83.1          & 72.3          & 75.9          \\ \bottomrule
				\end{tabular}\label{tav5-1} 
 \vspace{-2mm}
\end{table*}

\subsection{Effect of Weight Mask Block on ADPS} \label{sec4.D}

\begin{figure}
	\includegraphics[width=0.99\linewidth]{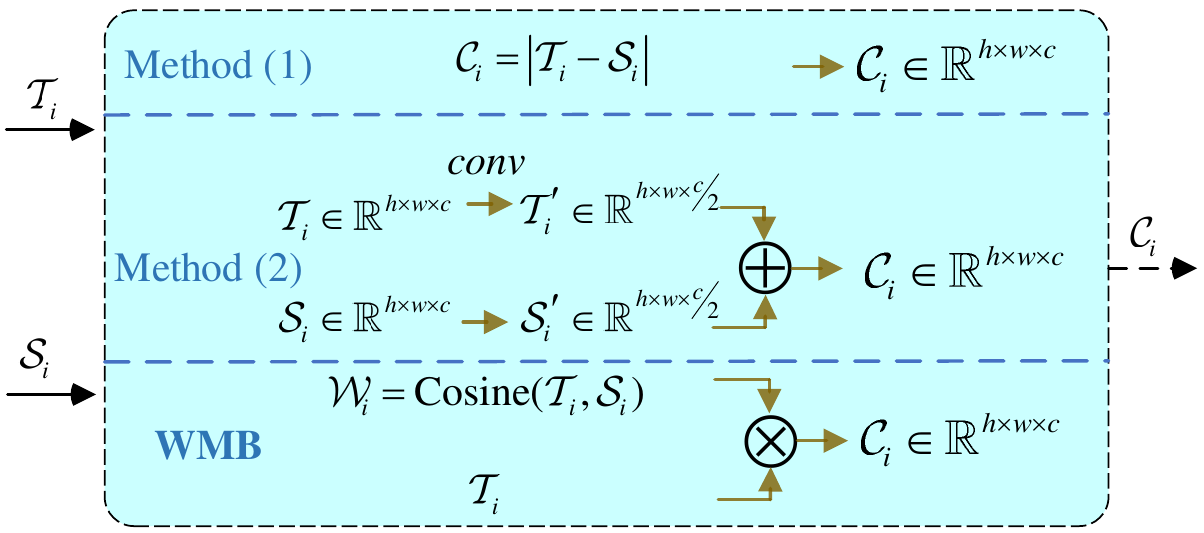}\vspace{-2mm}
	\caption{Illustration of the different ways to transfer distilled knowledge to $\mathcal{T}_i$.}
	\label{fig:figure7}
 \vspace{-2mm}
\end{figure}

We investigate three ways for combining distilled knowledge and pre-trained knowledge to obtain $C_i$, as illustrated in Figure~\ref{fig:figure7}.
\begin{itemize}
\item[$\bullet$]  Method (1): This method involves utilizing the differences in features between the teacher network and the student network as $C_i$, which is then directly input into PSM.

\item[$\bullet$]  Method (2): The features extracted from the student network and the teacher network are combined directly, where a $3\times 3$ convolution is employed to achieve dimensional reduction and fusion for obtaining $C_i$.

\item[$\bullet$]  WMB: The distilled knowledge $\mathcal{W}_i$ is leveraged in WMB to re-weight the original pre-trained representations, resulting in $C_i$.
\end{itemize}
The experimental results corresponding to the three ways are shown in Table~\ref{tav5}. Comparing Method (1) and Method (2) reveals that the anomaly detection capability of Method (1) significantly surpasses that of Method (2). 
This observation demonstrates that explicitly incorporating coarse localization information outperforms direct feature fusion. Accordingly, we design a more effective weighted method called WMB. WMB integrates not only the coarse localization information, $\mathcal{W}_i$, but also the multi-scale contextual information extracted from the pre-trained teacher network, $T_i$. WMB exhibits superior performance compared to Method (1), establishing it as the suitable module within the ADPS framework. Despite the simplicity of this weighted approach, this provides a solution for fusing distilled knowledge and pre-trained a priori knowledge for high-performance anomaly detection. Its advantage lies in ingeniously integrating distilled knowledge and pre-trained knowledge, thereby enabling PSM to derive fused features optimized for anomaly detection. 

Moreover, we investigate the effect of different methods for generating $\mathcal{W}_i$ in WMB. In addition to the cosine similarity loss (Cosine), we introduce Mean Squared Error (MSE) for generating $\mathcal{W}_i$. To ensure consistency of the distillation loss and the distilled target, the loss function of the distillation network is the same as the one used to generate $\mathcal{W}_i$ in WMB. For example, if MSE is used to generate a coarse localization mask, then the corresponding MSE loss is used as a distillation loss. Based on this premise, we introduce two comparison groups that employ both MSE and Cosine similarity losses. The experimental results are shown in table~\ref{tab6}. When using both MSE and Cosine similarity losses, the performance of anomaly detection is reduced. Employing the Cosine loss yields better anomaly detection and anomaly segmentation than using MSE. Interestingly, for ``textured" images, using MSE loss proves to be more effective.

\begin{table*}[t!]
	\centering
	\renewcommand\arraystretch{1.5}
	\caption{Anomaly detection results of different T-S feature fusion methods on the MVTec AD dataset. WMB achieves the best performance.}\vspace{-2mm}
			\setlength{\tabcolsep}{2.8mm}
   {
				\begin{tabular}{c|lll|lll|lll|lll}
					\toprule
					& \multicolumn{3}{c|}{$\mathcal{AUROC}_{cla}$}                     & \multicolumn{3}{c|}{$\mathcal{AUROC}_{seg}$}    &\multicolumn{3}{c|}{$\mathcal{PRO}_{seg}$}                 & \multicolumn{3}{c}{$\mathcal{AP}_{seg}$}                     \\ \cline{2-13}
					&Texture             & Object    & Mean             &Texture             & Object    & Mean           &Texture             & Object    & Mean  &Texture             & Object    & Mean \\ \hdashline
					Method(1) & \textbf{99.8} & \underline{99.2}          & \underline{82.3}          & \underline{94.6}          & \underline{95.7}        & \underline{69.1}         &\textbf{97.8}&\underline{88.9}&\underline{91.9} & \underline{96.4}          & \underline{96.9}          & \underline{73.5}          \\
					Method(2) & 99.6          & 99.0            & 81.7          & 94.5          & 92.8        & 66.8          &97.7&85.0&89.2& 96.1          & 94.9          & 71.8          \\ \hdashline
					\textbf{WMB}     &\underline{98.2} \textcolor{gray}{$\downarrow$}          & \textbf{99.5} \textcolor{orange}{$\uparrow$} & \textbf{83.1} \textcolor{orange}{$\uparrow$}& \textbf{97.4} \textcolor{orange}{$\uparrow$} & \textbf{97.0} \textcolor{orange}{$\uparrow$} & \textbf{74.3} \textcolor{orange}{$\uparrow$}& \textbf{97.8} \textcolor{orange}{$\uparrow$} &\textbf{90.6} \textcolor{orange}{$\uparrow$}& \textbf{93.0} \textcolor{orange}{$\uparrow$}& \textbf{97.7} \textcolor{orange}{$\uparrow$}& \textbf{97.8} \textcolor{orange}{$\uparrow$} & \textbf{77.2} \textcolor{orange}{$\uparrow$} \\ \bottomrule
				\end{tabular}\label{tav5}
	}\vspace{-2mm}
\end{table*}

\begin{table*}[t!]
	\centering
	\renewcommand\arraystretch{1.5}
	\caption{An ablation study on the selection of different knowledge distillation loss functions and different generation methods of $\mathcal{W}_i$.  }\vspace{-2mm}
			\setlength{\tabcolsep}{3mm}
   {
				
				\begin{tabular}{cc|cc|ccc|ccc|ccc}
					\hline
					\multicolumn{2}{c|}{Loss} & \multicolumn{2}{c|}{WMB} &         \multicolumn{3}{c|}{$\mathcal{AUROC}_{cla}$} & \multicolumn{3}{c|}{$\mathcal{AUROC}_{seg}$} & \multicolumn{3}{c}{$\mathcal{AP}_{seg}$} \\ \hline
					
					MSE           & Cosine           & MSE & Cosine &Texture             & Object    & Mean    &Texture             & Object    & Mean &Texture             & Object    & Mean     \\ \hline
					\ding{52}             & \ding{55}                 & \ding{52}    &  \ding{55}        & \textbf{99.4}    & 91.5   & 94.1   & \textbf{99.8}    & 92.4   & 94.5   & 77.3    & 56.6   & 63.5   \\
				\ding{55}  	& \ding{52}                & \ding{55}      & \ding{52}      & 98.3    & \textbf{97.4}   & \textbf{97.7}   & 99.5    & \textbf{97.0}     & \textbf{97.8}   & 83.1    & \textbf{74.3}   & \textbf{77.2}   \\ \hdashline
					\ding{52}              & \ding{52}                 & \ding{52}    &    \ding{55}      & 99.2    & 95.9   & 97.0     & 99.4    & 95.6   & 96.8   & \textbf{83.4}    & 71.0     & 75.2   \\
					\ding{52}              & \ding{52}                 & \ding{55}      & \ding{52}       & 99.2    & 96.6   & 97.5   & 99.5    & 96.0     & 97.2   & 83.3    & 71.8   & 75.6   \\ \hline
				\end{tabular}\label{tab6}
	}
 \vspace{-2mm}
\end{table*}

\subsection{Ablation Study on Framework of ADPS}
To investigate each component of the proposed ADPS framework, we compare ADPS with its variants on the MVTec-AD dataset.
\begin{itemize}
\item[$\bullet$] \emph{ W/O. PW}: The proposed ADPS without PSM and WMB. 
\item[$\bullet$] \emph{ W/O. T}: The proposed ADPS without the teacher network, \emph{i.e.,} direct training the student model and PSM.
\item[$\bullet$] \emph{ W/O. S}: The proposed ADPS removes the student network and directly uses a pre-trained teacher network. In this setting, only the parameters of the Post-Segmentation Module can be learned.
\end{itemize}
We show the variant models and their results in Table~\ref{fig:graph8}. Based on the experimental results, the following conclusions can be concluded:

\subsubsection{Pre-trained knowledge is essential for the teacher network}
The comparison results from \emph{W/O. T} and \emph{W/O. S} show a significant drop in anomaly detection and anomaly segmentation performance for multiple categories. Especially, the anomaly segmentation performance decreases by $10\%$ for $\mathcal{AP}_{seg}$. The experimental results confirm the effectiveness of prior knowledge from the pre-trained model for anomaly detection and emphasize the need for further exploration. Using only the teacher network is intuitively more effective than using only the student network, as the pre-trained model provides a powerful feature representation for the post-segmentation module. Another possible explanation is that after fixing the image encoder (teacher model), the PSM module is more capable of being trained more efficiently~\cite{DBLP:conf/eccv/QuWLGLRZW22}, yielding more accurate segmentation results.

\begin{figure}
	\centering
	\includegraphics[width=1.0\linewidth]{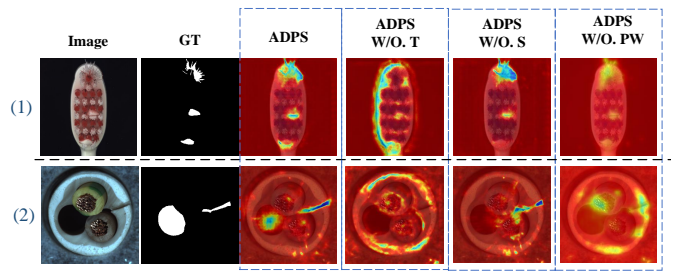}
	\caption{Anomaly heatmaps for four structures. ADPS captures more tiny anomaly regions with clearer and more detailed decision boundaries.}\vspace{-2mm}
	\label{fig:figure9}
 \vspace{-2mm}
\end{figure}

\subsubsection{Exploitation potential of the teacher network is significant} 
Comparing with ADPS, \emph{W/O. PW} shows a decrease of $7\%$ in $\mathcal{AUROC}_{cla}$ and $20\%$ in $\mathcal{AP}_{seg}$, highlighting the effectiveness and potential of the PSM and WMB modules in utilizing the features extracted by the teacher network. While the teacher network only serves as an extractor of reference features without leveraging its powerful feature representation capability, introducing WMB to utilize distilled knowledge and PSM to fully exploit the features extracted by the teacher model enhances the anomaly detection capability. Table~\ref{tav5} also demonstrates the significant impact of different methods that employ pre-trained knowledge and distilled knowledge on anomaly detection. The proposed WMB may not be the optimal scheme, and future research could explore methods that combine teacher networks and distillation strategies to further enhance anomaly detection ability.

\subsubsection{Asymmetric distillation and WMB contribute to the performance of ADPS.}  Comparing \emph{W/O. S} with ADPS, it can be observed that distilled knowledge obtained through asymmetric distillation significantly improves the performance of ADPS. 
However, without the WMB, the distilled knowledge cannot be effectively transferred to the teacher network. The corresponding results demonstrate that fully integrating pre-trained prior knowledge and distilled knowledge is effective in improving the performance of anomaly detection. The insight behind this is that prior knowledge provides powerful feature representation, while distilled knowledge enhances the discrimination between anomalies and normal instances.

\textbf{Feature Visualization.} We present visualizations in Figure~\ref{fig:figure9}, which includes two examples that compare anomaly segmentation results. Figure~\ref{fig:figure9} (1) and (2) illustrate the boundary between normal and abnormal features learned by \emph{W/O. S} is not accurate. 
Additionally, \emph{W/O. PW} shows the inability to derive the decision boundary between abnormal and normal from low-resolution feature maps, which leads to inefficient detection results. The limitations of \emph{W/O. S} in capturing differences in small anomalous regions are attributed to the lack of an asymmetrical distillation model that focuses on local regions. On the other hand, Figure~\ref{fig:figure9} demonstrates that $\mathcal{AP}_{seg}$ is more indicative of the performance of anomalous segmentation (\eg, although the segmentation results of \emph{W/O. PS} have many errors, $\mathcal{AUROC}_{seg}$ still maintains a high performance).

\textbf{Discussion.} Undoubtedly, the PSM plays a vital role, particularly in the anomaly segmentation capability of ADPS. It further confirms our hypothesis that solely relying on acquiring high-precision segmentation maps from low-resolution feature map differences is extremely challenging. Consequently, solving the anomaly segmentation challenge may have to rely on high-resolution images. Additionally, the approach employed by ADPS to exploit the features extracted by the pre-trained model is straightforward. We believe that more effective distilled knowledge can be obtained by leveraging the discrepancy between the learning capabilities of the teacher network and the student network.


\section{Conclusion}~\label{section5}

In this paper, we presented a novel knowledge distillation-based framework, ADPS, to address the main limitations of existing KDAD methods for image anomaly detection.
The proposed ADPS framework first establishes an asymmetric distillation paradigm that mitigates the student network's tendency to replicate feature representations from the teacher model, especially when presented with anomalous inputs. This asymmetric distillation paradigm also facilitates a greater emphasis on capturing local details within the distilled model.
Additionally, customized WMB and PSM were developed to fully incorporate the distilled posterior knowledge and pre-trained prior knowledge to produce high-quality segmentation maps with fine structures and clear boundaries.
Extensive experiments on three benchmarks demonstrate that our model performs favorably against state-of-the-art methods.
Importantly, our simple framework is modular, making it promising for serving as a strong baseline for future efficient KDAD research.
There's no denying that the dependence on simulated anomalous samples is the main limitation of ADPS, and we will give a better solution in future work.

\ifCLASSOPTIONcaptionsoff
  \newpage
\fi

{
\small
\bibliographystyle{IEEEtran}
\bibliography{egbib}

\begin{thebibliography}{10}
\providecommand{\url}[1]{#1}
\csname url@samestyle\endcsname
\providecommand{\newblock}{\relax}
\providecommand{\bibinfo}[2]{#2}
\providecommand{\BIBentrySTDinterwordspacing}{\spaceskip=0pt\relax}
\providecommand{\BIBentryALTinterwordstretchfactor}{4}
\providecommand{\BIBentryALTinterwordspacing}{\spaceskip=\fontdimen2\font plus
\BIBentryALTinterwordstretchfactor\fontdimen3\font minus
  \fontdimen4\font\relax}
\providecommand{\BIBforeignlanguage}[2]{{%
\expandafter\ifx\csname l@#1\endcsname\relax
\typeout{** WARNING: IEEEtran.bst: No hyphenation pattern has been}%
\typeout{** loaded for the language `#1'. Using the pattern for}%
\typeout{** the default language instead.}%
\else
\language=\csname l@#1\endcsname
\fi
#2}}
\providecommand{\BIBdecl}{\relax}
\BIBdecl

\bibitem{salehi2021unified}
M.~Salehi, H.~Mirzaei, D.~Hendrycks, Y.~Li, M.~H. Rohban, and M.~Sabokrou, ``A
  unified survey on anomaly, novelty, open-set, and out-of-distribution
  detection: Solutions and future challenges,'' \emph{arXiv preprint
  arXiv:2110.14051}, 2021.

\bibitem{pang2021deep}
G.~Pang, C.~Shen, L.~Cao, and A.~V.~D. Hengel, ``Deep learning for anomaly
  detection: A review,'' \emph{CSUR}, vol.~54, no.~2, pp. 1--38, 2021.

\bibitem{DBLP:journals/tnn/LiuHLTOL22}
J.~Liu, Z.~Hou, W.~Li, R.~Tao, D.~Orlando, and H.~Li, ``Multipixel anomaly
  detection with unknown patterns for hyperspectral imagery,'' \emph{IEEE
  TNNLS}, vol.~33, no.~10, pp. 5557--5567, 2022.

\bibitem{bergmann2019mvtec}
P.~Bergmann, M.~Fauser, D.~Sattlegger, and C.~Steger, ``Mvtec ad--a
  comprehensive real-world dataset for unsupervised anomaly detection,'' in
  \emph{CVPR}, 2019, pp. 9592--9600.

\bibitem{carrera2016defect}
D.~Carrera, F.~Manganini, G.~Boracchi, and E.~Lanzarone, ``Defect detection in
  sem images of nanofibrous materials,'' \emph{TII}, vol.~13, no.~2, pp.
  551--561, 2016.

\bibitem{defard2021padim}
T.~Defard, A.~Setkov, A.~Loesch, and R.~Audigier, ``Padim: a patch distribution
  modeling framework for anomaly detection and localization,'' in \emph{ICPR},
  2021, pp. 475--489.

\bibitem{fei2020attribute}
Y.~Fei, C.~Huang, C.~Jinkun, M.~Li, Y.~Zhang, and C.~Lu, ``Attribute
  restoration framework for anomaly detection,'' \emph{IEEE TMM}, 2020.

\bibitem{rudolph2021same}
M.~Rudolph, B.~Wandt, and B.~Rosenhahn, ``Same same but differnet:
  Semi-supervised defect detection with normalizing flows,'' 2021, pp.
  1907--1916.

\bibitem{salehi2021multiresolution}
M.~Salehi, N.~Sadjadi, S.~Baselizadeh, M.~H. Rohban, and H.~R. Rabiee,
  ``Multiresolution knowledge distillation for anomaly detection,'' in
  \emph{CVPR}, 2021, pp. 14\,902--14\,912.

\bibitem{deng2022anomaly}
H.~Deng and X.~Li, ``Anomaly detection via reverse distillation from one-class
  embedding,'' in \emph{CVPR}, 2022, pp. 9737--9746.

\bibitem{schlegl2019f}
T.~Schlegl, P.~Seeb{\"o}ck, S.~M. Waldstein, G.~Langs, and U.~Schmidt-Erfurth,
  ``f-anogan: Fast unsupervised anomaly detection with generative adversarial
  networks,'' \emph{MIA}, vol.~54, pp. 30--44, 2019.

\bibitem{baur2018deep}
C.~Baur, B.~Wiestler, S.~Albarqouni, and N.~Navab, ``Deep autoencoding models
  for unsupervised anomaly segmentation in brain mr images,'' in \emph{MICCAI
  workshop}, 2018, pp. 161--169.

\bibitem{DBLP:journals/tnn/JuniorY21}
F.~E.~F. Junior and G.~G. Yen, ``Automatic searching and pruning of deep neural
  networks for medical imaging diagnostic,'' \emph{IEEE TNNLS}, vol.~32,
  no.~12, pp. 5664--5674, 2021.

\bibitem{mejia2017pca}
A.~F. Mejia, M.~B. Nebel, A.~Eloyan, B.~Caffo, and M.~A. Lindquist, ``Pca
  leverage: outlier detection for high-dimensional functional magnetic
  resonance imaging data,'' \emph{Biostatistics}, vol.~18, no.~3, pp. 521--536,
  2017.

\bibitem{venkataramanan2020attention}
S.~Venkataramanan, K.-C. Peng, R.~V. Singh, and A.~Mahalanobis, ``Attention
  guided anomaly localization in images,'' in \emph{ECCV}, 2020, pp. 485--503.

\bibitem{vojir2021road}
T.~Vojir, T.~{\v{S}}ipka, R.~Aljundi, N.~Chumerin, D.~O. Reino, and J.~Matas,
  ``Road anomaly detection by partial image reconstruction with segmentation
  coupling,'' in \emph{ICCV}, 2021, pp. 15\,651--15\,660.

\bibitem{bogdoll2022anomaly}
D.~Bogdoll, M.~Nitsche, and J.~M. Z{\"o}llner, ``Anomaly detection in
  autonomous driving: A survey,'' in \emph{CVPR}, 2022, pp. 4488--4499.

\bibitem{gong2019memorizing}
D.~Gong, L.~Liu, V.~Le, B.~Saha, M.~R. Mansour, S.~Venkatesh, and A.~v.~d.
  Hengel, ``Memorizing normality to detect anomaly: Memory-augmented deep
  autoencoder for unsupervised anomaly detection,'' in \emph{ICCV}, 2019, pp.
  1705--1714.

\bibitem{park2020learning}
H.~Park, J.~Noh, and B.~Ham, ``Learning memory-guided normality for anomaly
  detection,'' in \emph{CVPR}, 2020, pp. 14\,372--14\,381.

\bibitem{salehi2020puzzle}
M.~Salehi, A.~Eftekhar, N.~Sadjadi, M.~H. Rohban, and H.~R. Rabiee,
  ``Puzzle-ae: Novelty detection in images through solving puzzles,''
  \emph{arXiv preprint arXiv:2008.12959}, 2020.

\bibitem{DBLP:journals/tnn/LiTLW22}
D.~Li, Q.~Tao, J.~Liu, and H.~Wang, ``Center-aware adversarial autoencoder for
  anomaly detection,'' \emph{IEEE TNNLS}, vol.~33, no.~6, pp. 2480--2493, 2022.

\bibitem{DBLP:journals/tnn/ZhouSZLZL22}
Y.~Zhou, X.~Song, Y.~Zhang, F.~Liu, C.~Zhu, and L.~Liu, ``Feature encoding with
  autoencoders for weakly supervised anomaly detection,'' \emph{IEEE TNNLS},
  vol.~33, no.~6, pp. 2454--2465, 2022.

\bibitem{bergmann2020uninformed}
P.~Bergmann, M.~Fauser, D.~Sattlegger, and C.~Steger, ``Uninformed students:
  Student-teacher anomaly detection with discriminative latent embeddings,'' in
  \emph{CVPR}, 2020, pp. 4183--4192.

\bibitem{QianWYHW22}
B.~Qian, Y.~Wang, H.~Yin, R.~Hong, and M.~Wang, ``Switchable online knowledge
  distillation,'' in \emph{{ECCV}}, vol. 13671, 2022, pp. 449--466.

\bibitem{wang2021student}
G.~Wang, S.~Han, E.~Ding, and D.~Huang, ``Student-teacher feature pyramid
  matching for unsupervised anomaly detection,'' \emph{arXiv preprint
  arXiv:2103.04257}, 2021.

\bibitem{bovzivc2021mixed}
J.~Bo{\v{z}}i{\v{c}}, D.~Tabernik, and D.~Sko{\v{c}}aj, ``Mixed supervision for
  surface-defect detection: From weakly to fully supervised learning,''
  \emph{CII}, vol. 129, p. 103459, 2021.

\bibitem{DBLP:conf/visapp/BergmannLFSS19}
P.~Bergmann, S.~L{\"o}we, M.~Fauser, D.~Sattlegger, and C.~Steger, ``Improving
  unsupervised defect segmentation by applying structural similarity to
  autoencoders,'' \emph{arXiv preprint arXiv:1807.02011}, 2018.

\bibitem{hou2021divide}
J.~Hou, Y.~Zhang, Q.~Zhong, D.~Xie, S.~Pu, and H.~Zhou, ``Divide-and-assemble:
  Learning block-wise memory for unsupervised anomaly detection,'' in
  \emph{ICCV}, 2021, pp. 8791--8800.

\bibitem{li2020superpixel}
Z.~Li, N.~Li, K.~Jiang, Z.~Ma, X.~Wei, X.~Hong, and Y.~Gong, ``Superpixel
  masking and inpainting for self-supervised anomaly detection,'' in
  \emph{BMVC}, 2020.

\bibitem{zavrtanik2021reconstruction}
V.~Zavrtanik, M.~Kristan, and D.~Sko{\v{c}}aj, ``Reconstruction by inpainting
  for visual anomaly detection,'' \emph{PR}, vol. 112, p. 107706, 2021.

\bibitem{you2022unified}
Z.~You, L.~Cui, Y.~Shen, K.~Yang, X.~Lu, Y.~Zheng, and X.~Le, ``A unified model
  for multi-class anomaly detection,'' in \emph{NIPS}, 2022.

\bibitem{wang2021cognitive}
T.~Wang, X.~Xu, F.~Shen, and Y.~Yang, ``A cognitive memory-augmented network
  for visual anomaly detection,'' \emph{IEEE JAS}, vol.~8, no.~7, pp.
  1296--1307, 2021.

\bibitem{xing2023visual}
P.~Xing and Z.~Li, ``Visual anomaly detection via partition memory bank module
  and error estimation,'' \emph{IEEE Transactions on Circuits and Systems for
  Video Technology}, 2023.

\bibitem{DBLP:conf/nips/GoodfellowPMXWOCB14}
I.~J. Goodfellow, J.~Pouget{-}Abadie, M.~Mirza, B.~Xu, D.~Warde{-}Farley,
  S.~Ozair, A.~C. Courville, and Y.~Bengio, ``Generative adversarial nets,'' in
  \emph{NeurIPS}, 2014, pp. 2672--2680.

\bibitem{schlegl2017unsupervised}
T.~Schlegl, P.~Seeb{\"o}ck, S.~M. Waldstein, U.~Schmidt-Erfurth, and G.~Langs,
  ``Unsupervised anomaly detection with generative adversarial networks to
  guide marker discovery,'' in \emph{ICIP in medical imaging}, 2017, pp.
  146--157.

\bibitem{akcay2018ganomaly}
S.~Akcay, A.~Atapour-Abarghouei, and T.~P. Breckon, ``Ganomaly: Semi-supervised
  anomaly detection via adversarial training,'' in \emph{ACCV}, 2018, pp.
  622--637.

\bibitem{anogan/SchleglSWSL17}
T.~Schlegl, P.~Seeb{\"{o}}ck, S.~M. Waldstein, U.~Schmidt{-}Erfurth, and
  G.~Langs, ``Unsupervised anomaly detection with generative adversarial
  networks to guide marker discovery,'' in \emph{Information Processing in
  Medical Imaging}, M.~Niethammer, M.~Styner, S.~R. Aylward, H.~Zhu, I.~Oguz,
  P.~Yap, and D.~Shen, Eds.

\bibitem{EfficientGAN/abs-1802-06222}
H.~Zenati, C.~S. Foo, B.~Lecouat, G.~Manek, and V.~R. Chandrasekhar,
  ``Efficient gan-based anomaly detection,'' \emph{arXiv preprint
  arXiv:1802.06222}, 2018.

\bibitem{zhang2020defgan}
D.~Zhang, S.~Gao, L.~Yu, G.~Kang, X.~Wei, and D.~Zhan, ``Defgan: Defect
  detection gans with latent space pitting for high-speed railway insulator,''
  \emph{TIM}, vol.~70, pp. 1--10, 2020.

\bibitem{chen2001one}
Y.~Chen, X.~S. Zhou, and T.~S. Huang, ``One-class svm for learning in image
  retrieval,'' in \emph{ICIP}, vol.~1, 2001, pp. 34--37.

\bibitem{ruff2018deep}
L.~Ruff, R.~Vandermeulen, N.~Goernitz, L.~Deecke, S.~A. Siddiqui, A.~Binder,
  E.~M{\"u}ller, and M.~Kloft, ``Deep one-class classification,'' in
  \emph{ICML}, 2018, pp. 4393--4402.

\bibitem{chalapathy2018anomaly}
R.~Chalapathy, A.~K. Menon, and S.~Chawla, ``Anomaly detection using one-class
  neural networks,'' \emph{arXiv preprint arXiv:1802.06360}, 2018.

\bibitem{yi2020patch}
J.~Yi and S.~Yoon, ``Patch svdd: Patch-level svdd for anomaly detection and
  segmentation,'' in \emph{ACCV}, 2020.

\bibitem{cohen2020sub}
N.~Cohen and Y.~Hoshen, ``Sub-image anomaly detection with deep pyramid
  correspondences,'' \emph{arXiv preprint arXiv:2005.02357}, 2020.

\bibitem{DBLP:conf/bmvc/ZagoruykoK16}
S.~Zagoruyko and N.~Komodakis, ``Wide residual networks,'' in \emph{BMVC},
  2016.

\bibitem{do2008multivariate}
C.~B. Do, ``The multivariate gaussian distribution,'' \emph{Section Notes,
  Lecture on Machine Learning}, vol. 229, 2008.

\bibitem{hazel2000multivariate}
G.~G. Hazel, ``Multivariate gaussian mrf for multispectral scene segmentation
  and anomaly detection,'' \emph{TGRS}, vol.~38, no.~3, pp. 1199--1211, 2000.

\bibitem{DBLP:journals/tnn/LiuXLZL23}
H.~Liu, X.~Xu, E.~Li, S.~Zhang, and X.~Li, ``Anomaly detection with
  representative neighbors,'' \emph{IEEE TNNLS}, vol.~34, no.~6, pp.
  2831--2841, 2023.

\bibitem{RudolphWRW22}
M.~Rudolph, T.~Wehrbein, B.~Rosenhahn, and B.~Wandt, ``Fully convolutional
  cross-scale-flows for image-based defect detection,'' pp. 1829--1838.

\bibitem{RothPZSBG22}
K.~Roth, L.~Pemula, J.~Zepeda, B.~Sch{\"{o}}lkopf, T.~Brox, and P.~V. Gehler,
  ``Towards total recall in industrial anomaly detection,'' in \emph{CVPR},
  2022, pp. 14\,298--14\,308.

\bibitem{DBLP:journals/corr/abs-2111-07677}
J.~Yu, Y.~Zheng, X.~Wang, W.~Li, Y.~Wu, R.~Zhao, and L.~Wu, ``Fastflow:
  Unsupervised anomaly detection and localization via 2d normalizing flows,''
  \emph{arXiv preprint arXiv:2111.07677}, 2021.

\bibitem{DBLP:journals/corr/abs-2301-12082}
G.~Xie, J.~Wang, J.~Liu, F.~Zheng, and Y.~Jin, ``Pushing the limits of fewshot
  anomaly detection in industry vision: Graphcore,'' \emph{arXiv preprint
  arXiv:2301.12082}, 2023.

\bibitem{DBLP:conf/miccai/RonnebergerFB15}
O.~Ronneberger, P.~Fischer, and T.~Brox, ``U-net: Convolutional networks for
  biomedical image segmentation,'' in \emph{MICCAI}, 2015, pp. 234--241.

\bibitem{zavrtanik2021draem}
V.~Zavrtanik, M.~Kristan, and D.~Sko{\v{c}}aj, ``Draem-a discriminatively
  trained reconstruction embedding for surface anomaly detection,'' in
  \emph{ICCV}, 2021, pp. 8330--8339.

\bibitem{lin2017focal}
T.-Y. Lin, P.~Goyal, R.~Girshick, K.~He, and P.~Doll{\'a}r, ``Focal loss for
  dense object detection,'' in \emph{ICCV}, 2017, pp. 2980--2988.

\bibitem{rippel2021modeling}
O.~Rippel, P.~Mertens, and D.~Merhof, ``Modeling the distribution of normal
  data in pre-trained deep features for anomaly detection,'' in \emph{ICPR},
  2021, pp. 6726--6733.

\bibitem{wu2021learning}
J.-C. Wu, D.-J. Chen, C.-S. Fuh, and T.-L. Liu, ``Learning unsupervised
  metaformer for anomaly detection,'' in \emph{ICCV}, 2021, pp. 4369--4378.

\bibitem{li2021cutpaste}
C.-L. Li, K.~Sohn, J.~Yoon, and T.~Pfister, ``Cutpaste: Self-supervised
  learning for anomaly detection and localization,'' in \emph{CVPR}, 2021, pp.
  9664--9674.

\bibitem{DBLP:journals/corr/abs-2110-03396}
J.~Song, K.~Kong, Y.-I. Park, S.-G. Kim, and S.-J. Kang, ``Anoseg: Anomaly
  segmentation network using self-supervised learning,'' \emph{arXiv preprint
  arXiv:2110.03396}, 2021.

\bibitem{tabernik2020segmentation}
D.~Tabernik, S.~{\v{S}}ela, J.~Skvar{\v{c}}, and D.~Sko{\v{c}}aj,
  ``Segmentation-based deep-learning approach for surface-defect detection,''
  \emph{JIM}, vol.~31, no.~3, pp. 759--776, 2020.

\bibitem{DBLP:journals/corr/KingmaB14}
D.~P. Kingma and J.~Ba, ``Adam: {A} method for stochastic optimization,'' in
  \emph{ICLR}, Y.~Bengio and Y.~LeCun, Eds., 2015.

\bibitem{DBLP:conf/cvpr/RisteaMINKMS22}
N.~Ristea, N.~Madan, R.~T. Ionescu, K.~Nasrollahi, F.~S. Khan, T.~B. Moeslund,
  and M.~Shah, ``Self-supervised predictive convolutional attentive block for
  anomaly detection,'' in \emph{CVPR}, 2022, pp. 13\,566--13\,576.

\bibitem{xing2022self}
P.~Xing, Y.~Sun, and Z.~Li, ``Self-supervised guided segmentation framework for
  unsupervised anomaly detection,'' \emph{arXiv preprint arXiv:2209.12440},
  2022.

\bibitem{kim2021semi}
J.-H. Kim, D.-H. Kim, S.~Yi, and T.~Lee, ``Semi-orthogonal embedding for
  efficient unsupervised anomaly segmentation,'' \emph{arXiv preprint
  arXiv:2105.14737}, 2021.

\bibitem{tao2022deep}
X.~Tao, X.~Gong, X.~Zhang, S.~Yan, and C.~Adak, ``Deep learning for
  unsupervised anomaly localization in industrial images: A survey,''
  \emph{TIM}, 2022.

\bibitem{DBLP:conf/eccv/QuWLGLRZW22}
M.~Qu, Y.~Wu, W.~Liu, Q.~Gong, X.~Liang, O.~Russakovsky, Y.~Zhao, and Y.~Wei,
  ``Siri: {A} simple selective retraining mechanism for transformer-based
  visual grounding,'' in \emph{ECCV}, ser. Lecture Notes in Computer Science,
  S.~Avidan, G.~J. Brostow, M.~Ciss{\'{e}}, G.~M. Farinella, and T.~Hassner,
  Eds., vol. 13695, pp. 546--562.

\end{thebibliography}
}

\vfill

\end{document}